\pdfoutput=1
\documentclass[11pt]{article}

\usepackage[final]{acl}
\usepackage{times}
\usepackage{latexsym}

\usepackage[T1]{fontenc}
\usepackage[utf8]{inputenc}

\usepackage{microtype}

\usepackage{inconsolata}

\usepackage{graphicx}
\usepackage{xspace, soul}
\usepackage{threeparttable}
\usepackage{booktabs}
\usepackage{multirow}
\usepackage{amsmath}
\usepackage{pifont}
\usepackage{amssymb,enumitem}
\usepackage{comment}
\usepackage{hyperref}
\usepackage{marvosym}
\usepackage{subfig}
\usepackage{caption}
\newcommand{\cmark}{\ding{51}}
\newcommand{\xmark}{\ding{55}}      
\soulregister{\cite}7
\soulregister{\ref}7
\soulregister{\st}7
\soulregister{\hl}7

\title{Captions Speak Louder than Images:
Generalizing Foundation Models \\
for E-commerce from High-quality Multimodal Instruction Data}

\author{
 \textbf{Xinyi Ling\textsuperscript{1}},
 \textbf{Hanwen Du\textsuperscript{1}},
 \textbf{Bo Peng\textsuperscript{1}},
 \textbf{Zhihui Zhu\textsuperscript{1}},
 \textbf{Xia Ning\textsuperscript{1 2 3 \Letter}}
\\
 \textsuperscript{1}Department of Computer Science and Engineering, The Ohio State University\\
 \textsuperscript{2}Translational Data Analytics Institute, The Ohio State University\\
 \textsuperscript{3}Department of Biomedical Informatics, The Ohio State University
\\
   \texttt{\{ling.303, du.1128, peng.707, zhu.3440, ning.104\}@osu.edu}
 }
\newcommand{\PRP}{\mbox{$\mathop{\mathtt{PRP}}\limits$}\xspace}
\newcommand{\CC}{\mbox{$\mathop{\mathtt{CC}}\limits$}\xspace}
\newcommand{\SA}{\mbox{$\mathop{\mathtt{SA}}\limits$}\xspace}
\newcommand{\SR}{\mbox{$\mathop{\mathtt{SR}}\limits$}\xspace}
\newcommand{\MPC}{\mbox{$\mathop{\mathtt{MPC}}\limits$}\xspace}
\newcommand{\PSI}{\mbox{$\mathop{\mathtt{PSI}}\limits$}\xspace}
\newcommand{\AP}{\mbox{$\mathop{\mathtt{AP}}\limits$}\xspace}

\newcommand{\dataset}{\mbox{$\mathop{\mathtt{MMECInstruct}}\limits$}\xspace}
\newcommand{\pipeline}{\mbox{$\mathop{\mathtt{CASLIE}}\limits$}\xspace}

\newcommand{\uniMMM}{\mbox{$\mathop{\mathtt{uniM^3}}\limits$}\xspace}
\newcommand{\majvote}{\mbox{$\mathop{\mathtt{MV}}\limits$}\xspace}
\newcommand{\allcap}{\mbox{$\mathop{\mathtt{UIA}}\limits$}\xspace}
\newcommand{\ECCC}{\mbox{$\mathop{\mathtt{EC^3}}\limits$}\xspace}
\newcommand{\CQE}{\mbox{$\mathop{\mathtt{CQE}}\limits$}\xspace}

\newcommand{\methodL}{\mbox{$\mathop{\mathtt{\uniMMM\text{-}L}}\limits$}\xspace}
\newcommand{\methodM}{\mbox{$\mathop{\mathtt{\uniMMM\text{-}M}}\limits$}\xspace}
\newcommand{\methodS}{\mbox{$\mathop{\mathtt{\uniMMM\text{-}S}}\limits$}\xspace}

\newcommand{\pipelineL}{\mbox{$\mathop{\mathtt{\pipeline\text{-}L}}\limits$}\xspace}
\newcommand{\pipelineM}{\mbox{$\mathop{\mathtt{\pipeline\text{-}M}}\limits$}\xspace}
\newcommand{\pipelineS}{\mbox{$\mathop{\mathtt{\pipeline\text{-}S}}\limits$}\xspace}

\begin{document}
\maketitle
% \def\thefootnote{*}\footnotetext{Equal contribution}

% \def\thefootnote{\arabic{footnote}}
%\hl{caution: when you refer to eCellM, make sure you do not say "in our previous work eCellM",  
%which will violate anonymity. }
%\xia{TODO: unify the style of highlighting important conclusions (bold, italic, et)}
%\xia{TODO: need to unify in the entire paper item vs product}.
%\xia{TODO: unify in-domain and IND, out-of-domain and OOD in experimental results}
%\xia{TODO: unify finetune vs fine-tune}
%\xia{TODO: at some places, mention \pipelineL, ..., \textit{ft-}}

\begin{abstract}
%
%\xia{need to mention we will release the dataset}

%Effectively harnessing multimodal data such as text and images for e-commerce exhibits strong promise.
%
%Given the recent surge of Large-Language Models
%(LLMs) on e-commerce tasks and their remarkable
%performance,
%
%Leveraging multimodal data to drive breakthroughs in e-commerce applications through multimodal foundation models (MFMs) is gaining increasing attention
%for improving product understanding and user experience.
% from the research community.
Multimodal foundation models (MFMs) have demonstrated strong capabilities in e-commerce by effectively leveraging multimodal data to enhance product understanding and user experience.
However, the development of e-commerce MFMs is hindered by two challenges: (1) the scarcity of large-scale, 
high-quality multimodal benchmark datasets; 
and (2) the lack of effective multimodal information integration methods in e-commerce.
To address these challenges,
we introduce \dataset, the first large-scale, high-quality multimodal instruction dataset
%for developing and evaluating foundation models 
% for e-commerce 
designed specifically for e-commerce MFMs.
\dataset comprises 75,000 samples covering 7 real-world e-commerce tasks, supporting both in-domain (IND) and out-of-domain (OOD) evaluations.
Leveraging \dataset, we develop \pipeline, a lightweight framework 
that enhances multimodal information understanding and integration for e-commerce.
%
% Leveraging \dataset, 
% we fine-tune a series of 
% e-commerce MFMs within \pipeline, denoted as \pipeline models.
%
Our comprehensive evaluation demonstrates that
\dataset endows \pipeline with advanced capability %(6.4\% overall improvement over baselines) 
and strong generalizability %(3.9\% improvement) 
in e-commerce applications.
% \pipeline models substantially outperform 5 categories of advanced baseline models in the in-domain evaluation of \dataset, and show strong generalizability
% to out-of-domain settings.
%
\dataset and \pipeline models are publicly accessible
through \url{https://ninglab.github.io/CASLIE/}.
% \href{https://ninglab.github.io/CASLIE/}{https://ninglab.github.io/CASLIE/}.
%
%\xia{too long, need to reduce}

\end{abstract}

%%%%%%%%%%%%%%%%%%%%%%%%%%%%%%%%%%%%%%%%
\section{Introduction}
%%%%%%%%%%%%%%%%%%%%%%%%%%%%%%%%%%%%%%%%

% challenge: 
% 1. no benchmark dataset
% 2. integrate multimodal data
% current methods do not consider e-commerce data cricristic / feature (low quality), information not fully match what user want to see / things the user want to see is not obvious/ only part is informative. 
% how to capture this info flexible robust
% do not know which info of text/image is useful, which align, which unaligned, which can be connected/use together
% current CLIP align data, no need to align(convey different information) cannot deal with
% 3. generalist foundation model. with integrated data how to train various models.
% use data from various tasks together, emphasize in multimodal not e-commerce
% important: multimodal, data type

Multimodal data, encompassing diverse modes and types of information
such as text and images, is ubiquitous and essential for many real-world applications~\cite{antol2015vqa,MISSRec,mu2024robocodex,chen-etal-2021-multimodal-item}. In e-commerce, multimodal data is especially important: product content typically combines visual and textual information, and user interactions involve diverse data types across multiple modalities.
Effectively harnessing multimodal data for e-commerce
%via combining product specifics, textual reviews, visual content, etc., 
exhibits strong promise  
to allow for a more comprehensive depiction of product attributes
and uncover deeper insights into customer preferences, 
which single-modal data alone may not suffice~\cite{MISSRec,peng2023multi}. 
%
%is crucial 
%for building robust systems capable of handling e-commerce tasks such as recommendations. 
%
With the recent surge of Large-Language Models (LLMs) on e-commerce tasks and their 
remarkable performance~\cite{peng2024ecellm,li2024ecomgpt,shi2023llama}, 
%\hl{it is reasonable to expect that multimodal data can drive new breakthroughs in e-commerce applications 
%through the use of LLMs (i.e., unimodal foundation models) or Multimodal Foundation Models (MFMs).}
multimodal data are expected to drive new breakthroughs in e-commerce applications, 
together with the development of Multimodal Foundation Models (MFMs).
%or other foundation models.  

\begin{figure*}
    \centering
    \subfloat[\dataset Overview\label{fig:dataset}]{\includegraphics[width=0.72\textwidth]{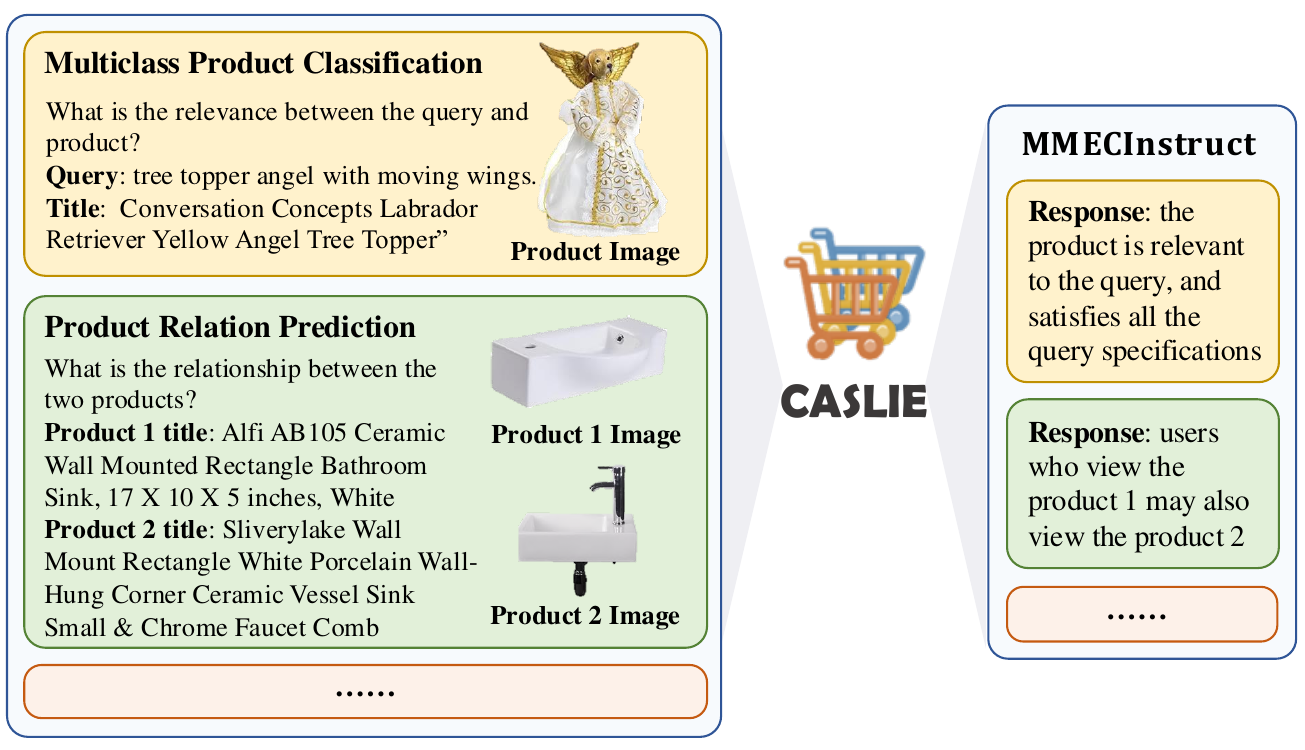}} \hfill
    \subfloat[Workflow of \pipeline \label{fig:caslie}]{\includegraphics[width=0.23\textwidth]{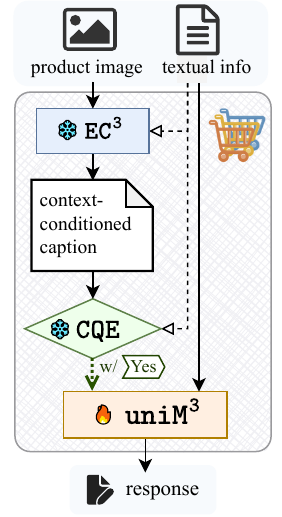}}
    \vspace{-5pt}
    \caption{\dataset and \pipeline overview}
    \label{fig:overview}
    \vspace{-12pt}
\end{figure*}

% \begin{figure*}
% \vspace{-3pt}
% 	\centering
% 	\begin{minipage}[c]{0.715\textwidth}
%          \includegraphics[width=0.95\linewidth]{figure/dataset2.pdf}
%          \caption{\dataset Dataset Overview}
%          \label{fig:dataset}
% 	\end{minipage}
% 	\hfill
% 	\begin{minipage}[c]{0.25\textwidth}
% 	\centering
%          \includegraphics[width=0.9\linewidth]{figure/caslie.pdf}
%          \caption{\pipeline Overview}
%          \label{fig:caslie}
% 	\end{minipage}
% 	\vspace{-11pt}
% \end{figure*}

However, despite the richness of multimodal e-commerce data, there are significant challenges that hinder its optimal use 
by foundation models~\cite{MISSRec,liu2023multimodal}:
\textbf{(1) Scarcity of large-scale, high-quality multimodal benchmark datasets for a large variety of e-commerce applications.} 
It is highly nontrivial to curate such a dataset due to the complexity of the data processing involved
(e.g., selecting products that possess rich, high-quality data across all modalities). 
While initiatives for unimodal e-commerce benchmark datasets for LLMs have been undertaken~\cite{peng2024ecellm,li2024ecomgpt,shi2023llama}, to the best of our knowledge, no such multimodal counterparts exist. 
\textbf{(2) Lack of effective multimodal information integration methods for e-commerce tasks.} 
Current LLM-based e-commerce models~\cite{peng2024ecellm,li2024ecomgpt} often focus predominantly on one modality, typically text.  
%and underutilize visual information and other signals, significantly impeding progress in the e-commerce domain.
Existing multimodal approaches~\cite{chia2022fashionclip, yu2022commercemm} attempt to map different modalities into a shared latent space, following the CLIP paradigm~\cite{radford2021clip} developed from the computer vision domain. 
However, this alignment-based strategy overlooks 
key challenges unique to e-commerce. 

First, multimodal information often complements rather than aligns~\cite{lin2025contrastive,dufumier2025what,baldrati2022effective}, while alignment is a core assumption in CLIP. 
For instance, an image of a large shampoo bottle conveys information about its bottle size but not its fragrance, while user reviews may praise its fragrance. Thus, image and user reviews are complementary 
to each other. % but complain about the bottle size. 
Second, the relevance of visual information is highly context-dependent: the same image feature may be crucial in one product category but irrelevant in another~\cite{li2014impact,gu2024exploring}.
%
% This strategy ignores the unique characteristics of e-commerce data
% \xia{
% : e-commerce multimodal data do not always align 
% -- a key assumption of CLIP-based models, but complement.
% } \hl{check if this is accurate}
% %
% For example, an image of a big bottle of shampoo does not 
% contain information on its scent, while user reviews praise the scent but complain about the bottle size -- information 
% alignment does not always occur.   
%
%\xia{add one more point here: useful image information is context-specific -- need to highlight these challenges; let's discuss this}

To address these challenges, 
we introduce \underline{\dataset}, the first-ever, large-scale, and high-quality multimodal instruction dataset 
designed specifically for e-commerce applications.
As shown in Figure~\ref{fig:dataset}, \dataset consists of 75,000 samples spanning 7 widely-performed real-world e-commerce tasks.
Each data sample includes an instruction, one or multiple images, a textual input, and an expected response, enabling the development and evaluation of e-commerce foundation models. 
\dataset is carefully curated to support a broad range of experimental settings, including in-domain (IND) evaluation for all 7 tasks, out-of-domain (OOD) evaluation (i.e., evaluation task on products of new category not included in the training set) for 5 tasks, and task-specific studies, ensuring robustness in real-world scenarios.
We perform rigorous processing to ensure the high quality of the \dataset.
%\xia{need to rephrase this paragraph. cannot use the same language as in eCellM. @Hanwen can help.}
%

% To better integrate visual and textual information for target tasks,
Leveraging \dataset,
we develop \underline{\pipeline} -- \textbf{CA}ptions \textbf{S}peak \textbf{L}ouder than \textbf{I}mag\textbf{E}s, 
a simple, lightweight, yet effective learning framework for e-commerce MFMs, 
which integrates text and images for e-commerce tasks. Figure \ref{fig:caslie} shows the workflow of \pipeline.
%
%We also develop \pipeline -- \underline{ca}ptions \underline{s}peak \underline{l}ouder than \underline{i}mag\underline{e}s, a framework 
\pipeline comprises three modules: 
\textbf{(1)} a context-conditioned caption generation module, denoted as \ECCC, 
that translates images into captions conditioned on given context,
% generating context-aware captions, textual representations
% filtering out ineffectual visual data, 
% and strategically integrating multimodal information.
%
\textbf{(2)} a caption quality evaluation module, denoted as \CQE, that excludes ineffectual visual information,
and 
\textbf{(3)} a modality information fusion module, denoted as \uniMMM, that seamless integrates visual and textual information for downstream tasks.
\pipeline processes images in a way that adapts to product-specific contexts, 
generating high-quality captions that bridge visual and text in a context-aware way, 
%
% \xia{\pipeline is foundamentally different -- why? need 1 or 2 sentences here talking about the novelty}
making it fundamentally different from previous work~\cite{chia2022fashionclip, liu2023visual}.

Existing MFMs typically embed and align visual and textual inputs using context-agnostic fusion techniques~\citep{li2024llava-next-interleave}. 
However, they often fail to distinguish helpful content from noise in images, resulting in suboptimal multimodal representations for e-commerce applications. 
Different from these models, \pipeline offers a simple, light-weight, \mbox{training-free} yet effective fusion framework, 
enabling a unified view of multimodal data for e-commerce tasks.
Another advantage of {\pipeline} is its plug-and-play design: all its modules can be easily reimplemented when newer and more advanced models become available, allowing for seamless integration of the most suitable options.
%
% \xy{Unlike prior work that relies on static or heuristic multimodal fusion, 
% \pipeline processes images through context-conditioned captioning. 
% This approach enables it to selectively extract and integrate only the most task-relevant visual information, 
% preserving both interpretability and 
% the benefits of multimodal learning in a task-adaptive manner.}
%
% This lightweight, modular approach enables scalable deployment while 
%
Our experiments show that 
\pipeline is significantly empowered with \dataset
% \dataset significantly enhances multimodal learning, enabling \pipeline 
to outperform state-of-the-art baselines across multiple e-commerce tasks.
We make \dataset publicly available at {\href{https://ninglab.github.io/CASLIE/}{https://ninglab.github.io/CASLIE/}}
to facilitate further research in multimodal learning for e-commerce. 

\begin{table}
%\vspace{-3pt}
  \centering
  %\vspace{-6pt}
  \setlength{\tabcolsep}{1pt}
  \begin{footnotesize}
  \begin{threeparttable}
      \begin{tabular}{
ccrcc
      }
      \toprule
       \multicolumn{1}{c}{\textbf{Mod.}} & \multicolumn{1}{c}{\textbf{Dataset}} & \multicolumn{1}{c}{\textbf{Size}}  & \multicolumn{1}{c}{\textbf{Div.}}  & \multicolumn{1}{c}{\textbf{Ins.}} \\
	 \midrule
       \multirow{6}{*}{Text}
       & Amazon-M2~\cite{jin2024amazonm2} & 3.6M & 	 \xmark & \xmark \\
       & Shopping Queries~\cite{reddy2022shopping} & 130K & $\circ$ & \xmark\\
       & EcomInstruct~\cite{li2024ecomgpt} & 2.6M & \cmark & \cmark \\
       & ECInstruct~\cite{peng2024ecellm} & 116K & \cmark & \cmark \\
       & Shopping MMLU~\cite{jin2024shoppingmmlu} & 11K & \cmark & \xmark \\
       & AmazonQA~\citep{gupta2019amazonqa} & 924K & $\circ$ & \xmark \\
       \midrule
       Text \&
       & MEP-3M$^\ast$~\cite{liu2023mep} & 3M & \xmark & \xmark \\
       Image& \dataset (ours) & 75K & \cmark & \cmark \\
      \bottomrule
      \end{tabular}
      % \begin{tablenotes}[normal,flushleft]
      % \begin{footnotesize}
      % \item 
      % $^\circ$The datasets only contain query/QA-related tasks. 
      % $^\ast$MEP-3M is composed of product metainformation, which lacks structured instructions for downstream applications.
      % % Amazon-M2 dataset is comprised of users' interaction sessions and product information. 
      % \par
      % \end{footnotesize}
      % \end{tablenotes}
  % \vspace{-13pt}
  \end{threeparttable}
  \end{footnotesize}
  %\vspace{-2pt}
  \caption{Comparison with existing e-commerce dataset. ``Mod.'' denotes the type of modalit(ies) in the dataset. ``Size'' denotes the number of samples in each dataset. ``Div.'' denotes whether the dataset contains diverse tasks. ``Ins.'' denotes whether the dataset contains instructions for LLM finetuning. 
    $^\ast$MEP-3M is composed of product meta information, lacking structured formulation for downstream applications.
  $^\circ$The datasets only contain query/QA-related tasks. 
  }
  \label{tbl:data_comp}
  % \vspace{-10pt}
\end{table}
% \label{tbl:data_comp}

%%%%%%%%%%%%%%%%%%%%%%%%%%%%%%%%
\section{Related Work}
%%%%%%%%%%%%%%%%%%%%%%%%%%%%%%%%

%%=============================================================
%\subsection{Vision and Language Instruction Tuning}
%%=============================================================
%
%Fine-tuning MFMs on paired instruction-output data~\cite{ouyang2022training,peng2023instruction,liu2023visual} 
%(i.e., instruction tuning) enables MFMs to follow the users' instructions and boosts the generalizability of MFMs~\cite{wei2022finetuned,chung2024scaling}.
%%
%Considering the importance of high-quality datasets for instruction tuning~\cite{chung2024scaling}, various instruction tuning datasets~\cite{wang-etal-2023-self-instruct,liu2023mitigating,dai2023instruct} are collected to fine-tune MFMs on general-purpose multi-modal tasks, such as multi-modal question answering~\cite{liu2023visual,liu2024improved} and image-text retrieval~\cite{li2023blip2}.
%%
%\emph{Different from existing instruction datasets for general-purpose MFMs, we curate a novel, high-quality multi-modal instruction dataset \dataset for e-commerce.} 
%%\bo{@Hanwen we may also want to mention different from existing instruction tuning data for MFM, we deliberately exclude less informative images.}
%
%%\xia{need to discuss this section: what is the point here -- instruction tuning (the method) or the data?}

%++++++++++++++++++++++++++++++++++++++++++++
\paragraph{E-commerce Benchmark}
%++++++++++++++++++++++++++++++++++++++++++++
Developing MFMs for e-commerce 
requires high-quality datasets that integrate multimodal information. 
Several existing datasets focus on text-based e-commerce tasks, such as EcomInstruct~\cite{li2024ecomgpt} and ECInstruct~\cite{peng2024ecellm}, 
which provide instruction-based learning resources but lack image data, 
limiting their applicability for multimodal learning. 
Other datasets, such as Amazon-M2~\cite{jin2024amazonm2} and the Shopping Query Dataset~\cite{reddy2022shopping}, 
contain large-scale e-commerce interactions but primarily focus on user behavior and query-related tasks without multimodal coverage. 
While MEP-3M~\cite{liu2023mep} incorporates both text and image modalities, it lacks structured instructions, 
making it less suitable for fine-tuning instruction-following multimodal models. 
In contrast, \textit{\dataset is the first multimodal instruction dataset for e-commerce, offering task-specific, high-quality image-text pairs across seven diverse e-commerce applications.} 
By addressing these gaps, \dataset establishes a new benchmark for multimodal e-commerce research, enabling robust evaluation and generalization of foundation models.

\vspace{-2pt}
%++++++++++++++++++++++++++++++++++++++++++++
\paragraph{Multimodal Learning for E-commerce}
%++++++++++++++++++++++++++++++++++++++++++++
\vspace{-1pt}
In recent years, remarkable advancements in multimodal learning~\cite{radford2021clip,li2021align,alayrac2022flamingo,stevens2024bioclip} have enabled 
significant process in integrating vision and language into e-commerce models. 
For example, CommerceMM~\cite{yu2022commercemm} learns multimodal representations for various e-commerce tasks by aligning paired data from different modalities via contrastive learning.
ECLIP~\cite{jin2023learning} and FashionCLIP~\cite{chia2022fashionclip} introduce CLIP~\cite{radford2021clip}-based contrastive pre-training frameworks to learn 
multimodal e-commerce data representations transferable to downstream tasks.
%
%\hl{However, aligning multimodal representations is not context-conditioned, that is, 
%the model learns unified representations even when the text and image modalities convey different information} \xia{this is not what we mean by context-conditioned}~\cite{liang2023factorized,dufumier2024align}.
%
However, CLIP-based models generate image representations from the entire image in a context-free manner, making it difficult to emphasize specific image details conditioned on the given context.
In contrast, \textit{\pipeline generates context-conditioned textual representations for images (e.g., captions), highlighting different details depending on the context.}
Additionally, \pipeline leverages the world knowledge in MFMs to generate captions, enriching captions with additional information pertinent to target tasks.

%
%\emph{
%\hl{In this paper, the proposed {\pipeline} can effectively leverage the world knowledge in MFMs to generate context-conditioned textual representations for images.} \xia{this is not the point and 
%related to the above literature. revise}}
%\bo{@Hanwen I think the major point here is to generate context-conditioned textual representations for images and \pipeline integrate the world knowledge in MFMs to enhance e-commerce tasks.} \xia{agree}
%
%\subsection{Modality Alignment}
%
%In recent years, significant advancements have been made in multimodal learning to seek the unified integration of the vision and language. CLIP~\cite{radford2021clip} innovatively learns the image-text pair in a contrastive pattern, which involves the inductive bias that the pair is inherently aligned. The successive works~\cite{chia2022fashionclip, stevens2024bioclip} adapt the CLIP-based method in different domains. However, in most real-world scenarios, multimodal information is not inherently aligned, such as the e-commerce realm. With the emergence of rapidly developed LLMs, sequential models are becoming popular in solving multimodal problems. LLaVA series works~\cite{liu2023improvedllava, li2024llava-next-interleave} delve deep into incorporating vision embedding with the LLMs. The natural comprehensive capability of LLMs enables the model to understand both the image and text. 

%%%%%%%%%%%%%%%%%%%%%%%%%%%%%%%%

%%%%%%%%%%%%%%%%%%%%%%%%%%%%%%%%
\vspace{-1pt}
%%%%%%%%%%%%%%%%%%%%%%%%%%%%%%%%%%%%%%%%%%%%%%%%%
\section[]{\dataset Dataset}
\label{sec:dataset}
%%%%%%%%%%%%%%%%%%%%%%%%%%%%%%%%%%%%%%%%%%%%%%%%%
\vspace{-1pt}
To advance multimodal learning in e-commerce, 
we introduce \dataset, a multimodal instruction dataset designed to adapt general-purpose MFMs
% \hl{LLMs} \xia{multimodal foundation models} 
for e-commerce.
\dataset is constructed under three principles: 
\textbf{(1)} \textbf{Multimodalilty}: 
Unlike text-only datasets (e.g., EcomInstruct~\cite{li2024ecomgpt} and Shopping MMLU~\cite{jin2024shoppingmmlu}), 
\dataset contains both visual and textual content for each product in various e-commerce tasks,
% \hl{enabling LLMs to achieve a more comprehensive understanding of the context from the multimodal perspective}
enabling comprehensive multimodal learning of foundation models.
% allowing foundation models to incorporate such multimodal data and enhance e-commerce performance.
%
% \hl{Multimodal data is critical for enabling LLMs to achieve a more comprehensive understanding of context}~\cite{achiam2023gpt}.\hanwen{this sentence does not transition smoothly.}
%
\textbf{(2) Broad coverage}: 
\dataset comprises seven diverse and realistic tasks to enable versatile e-commerce modeling 
and benchmarking~\cite{peng2024ecellm,jin2024shoppingmmlu,jin2024amazonm2}. %\xia{comparison with literature?}
\textbf{(3) High quality}:
The dataset is carefully curated through rigorous validation processes to ensure both accuracy and reliability.
As demonstrated in the literature~\cite{hoffmann2022training,gadre2024datacomp}, \mbox{high-quality} \mbox{instruction-tuning} data plays a pivotal role in building powerful foundation models.
Figure~\ref{fig:dataset} presents the overview of \dataset, and Table~\ref{tbl:data_comp} summarizes related e-commerce datasets. More information about \dataset dataset can be found in Appendix~\ref{sec:appendix:dataset}.
\textbf{\textit{To the best of our knowledge, MMECInstruct is the first of its kind.}} %\xia{fix this, \dataset should be bold and itatics in this case}

%\xia{this paragraph is not necessary; combine it with the previous paragraph. }
%Table~\ref{tbl:data_comp} summarized related e-commerce datasets.
%
%\xia{too much and over-claim}
%By providing high-quality, instruction-driven, multimodal data,
%\dataset sets a new standard for training and evaluating foundation models in e-commerce. 
%The dataset is publicly available to encourage further research and innovation in multimodal learning for real-world applications.
% 

%Due to the lack of multimodal e-commerce instruction datasets, we introduce the first multimodal e-commerce instruction dataset \dataset, a meticulously curated benchmark to endow LLMs with the capability to comprehensively understand both products and users from multimodal views. (Figure). \dataset is comprised of 75K instances from 7 real-world applications. \dataset aims to assess the model's perception and comprehensive abilities in handling multimodal information in the e-commerce domain, which requires more than just basic vision-language understanding, demanding approaches to incorporate domain-specific knowledge.

\vspace{-3pt}
%++++++++++++++++++++++++++++++++++++++++++++++++
\subsection{E-commerce Tasks}
\label{sec:dataset:tasks}
%++++++++++++++++++++++++++++++++++++++++++++++++

%Following ECInstruct~\cite{peng2024ecellm}, 
%\dataset comprises 7 widely-performed real-world tasks constructed from real-world data, which are ubiquitous and essential in the e-commerce domain. 
%
\begin{table*}
% \vspace{-5pt}
  \centering
  \begin{footnotesize}
  \begin{threeparttable}
      \begin{tabular}{
	@{\hspace{2pt}}l@{\hspace{4pt}}
	@{\hspace{4pt}}p{0.35\textwidth}@{\hspace{2pt}}
	@{\hspace{2pt}}p{0.12\textwidth}@{\hspace{1pt}}
	@{\hspace{0pt}}p{0.15\textwidth}@{\hspace{0pt}}
	@{\hspace{2pt}}p{0.3\textwidth}@{\hspace{0pt}}
      }
      \toprule
       \textbf{Task} & \textbf{Definition}  & \textbf{Type}  & \textbf{Primary Metrics}  & \textbf{Data Source}  \\
	 \midrule
       \multirow{2}{*}{\AP} & Predict if the product-related question is & Binary & \multirow{2}{*}{F1 score} & AmazonQA \\
       & answerable based on the product information. & classification  & & \cite{gupta2019amazonqa} \\
       \cmidrule(lr){1-5}
       \multirow{2}{*}{\CC} & Retrieve the category of the product based & \multirow{2}{*}{Retrieval} & \multirow{2}{*}{Recall@1} & MAVE~\cite{yang2022mave}, Amazon \\
       & on the product information. & & & Review 2023~\cite{hou2024bridging} \\
       \cmidrule(lr){1-5}
       \multirow{2}{*}{\PRP} & Identify the relationship between two product & Multi-class & \multirow{2}{*}{Macro F1 score} & Amazon Review 2023 \\
       & from \textit{``also buy", ``also view"}, and \textit{``similar"}. & classification & & \cite{hou2024bridging} \\
       \cmidrule(lr){1-5}
       \multirow{2}{*}{\PSI} & Predict if the product can serve as a & Binary & \multirow{2}{*}{F1 score} & Shopping Query Dataset\\
       & functional substitute for the user’s query. & classification && \cite{reddy2022shopping} \\
       \cmidrule(lr){1-5}
       \multirow{2}{*}{\MPC} & Given a query and product information, pre- & Multi-class & \multirow{2}{*}{Accuracy} & Shopping Query Dataset \\
       & dict relevance between the query and product. & classification && \cite{reddy2022shopping} \\
       \cmidrule(lr){1-5}
       \multirow{2}{*}{\SA} & Identify the sentiment user expressed based & Multi-class & \multirow{2}{*}{Macro F1 score} & Amazon Review 2023 \\
       & on the product review text and review image. & classification && \cite{hou2024bridging} \\
       \cmidrule(lr){1-5}
       \multirow{2}{*}{\SR} & Predict the next product that user would be & \multirow{2}{*}{Retrieval} & \multirow{2}{*}{Recall@1} & Amazon Review 2023 \\
       & interested in based on user’s purchase history. &&& \cite{hou2024bridging} \\
      \bottomrule
      \end{tabular}
      % \begin{tablenotes}[normal,flushleft]
      % \begin{footnotesize}
      % \item
      % % 
      % \par
      % \end{footnotesize}
      % \end{tablenotes}
  \end{threeparttable}
  \end{footnotesize}
  \vspace{-2pt}
  \caption{Tasks in \dataset dataset}
  \label{tbl:tasks}
  \vspace{-10pt}
\end{table*}
% \label{tbl:tasks}

%Points: 1) list all tasks; 2) highlight these tasks are real-world e-commerce tasks.
In line with prior works~\cite{yue2023mammoth, fang2024molinstructions, peng2024ecellm}, \dataset comprises 7 widely-performed \mbox{real-world} e-commerce tasks with real-world data extracted from e-commerce platforms:
% \xia{more citations from other domains on how to construct benchmarking datasets}:
\textbf{(1)} answerability prediction (\AP)~\cite{gupta2019amazonqa}, 
\textbf{(2)} category classification (\CC)~\cite{yang2022mave,chen-etal-2021-multimodal-item}, 
\textbf{(3)} product relation prediction (\PRP)~\cite{ni2019justifying, xu2020prp}, 
\textbf{(4)} product substitute identification (\PSI)~\cite{reddy2022shopping}, 
\textbf{(5)} multi-class product classification  (\MPC)~\cite{reddy2022shopping},
\textbf{(6)} sentiment analysis (\SA)~\cite{ wankhade2022sa,daza2024sentiment}, and 
\textbf{(7)} sequential recommendation (\SR)~\cite{li2023recformer,hou2024bridging,petrov2023gsasrec}.
These tasks are designed %\st{based on coverage of} \xia{to cover}
to cover key %\st{e-commerce}
functions in modern e-commerce platforms, including search, recommendation, QA, and sentiment
analysis. 
% \xy{by retrieval or classification prediction}. %\xia{
 %}\st{, to evaluate the method's capability to handle user's questions, queries, and recommendation requests in real-world scenarios.} \xia{this is tedious and redundant. revise}
%
Detailed information about all the e-commerce tasks is presented in Table~\ref{tbl:tasks}.
\subsection{Vision-language Data}
\label{sec:dataset:curation}
%++++++++++++++++++++++++++++++++++++++++++++++++
% \vspace{-1pt}
%Points: 1) what images are included in each task; 2) \dataset fundamentally different from existing work on including vision-language data

%\st{We construct {\dataset} from diverse, real-world e-commerce data sources listed in Table~\ref{tbl:tasks}, carefully curating multimodal content to ensure rich, informative, and high-quality samples.}

Different from existing datasets with text-only instructions~\cite{peng2024ecellm}, \dataset includes both visual and textual content for each item.
Particularly, 
the dataset includes \textbf{(1)}  %\xia{if (1), (2), (3), ... are bold, they should be bold all the time in the entire paper -- please unify the style!} 
product images and user review images as visual information, \textbf{(2)} product titles, product categories, product brands, user queries, user reviews, and user questions as textual content, \textbf{(3)} %\xy{human-generated}
human-designed structured instructions tailored to real-world scenarios for each task, 
% \st{allowing models to learn from practical e-commerce applications,} 
and \textbf{(4)} ground-truth response to each sample.
% \xia{need to specify what this is?}
% each sample in \AP contains product reviews provided by the users as text and product images; in the \SA task, each sample contains review text and a review image provided by the user; in the \SR
% . In all other tasks, 
% %
% In \CC, each sample has XXX.
%
% \bo{@Xinyi please complete this section}
%
% More statistics of \dataset can be found in Appendix~\ref{sec:appendix:dataset}.
% Specific samples are described in Appendix~\ref{sec:appendix:case}.
%
The multimodal e-commerce data is enriched with synergistic visual and textual inputs, providing a basis for developing and evaluating models on a range of multimodal e-commerce tasks.
% \xia{rephrase -- you are praising your own stuff too much and in the same language}
%ultimately improving models' effectiveness on multimodal e-commerce tasks.
% \hl{providing LLMs with a more comprehensive understanding of the context from a multimodal perspective.} \xia{@Bo: revise, not accurate.}

% \xy{Providing LLMs with rich information from different modalities allows for a deeper understanding of the task, as it combines both visual and textual data in a meaningful way.}

\begin{comment}
%++++++++++++++++++++++++++++++++++++++++++++++++
\subsection{Data Curation}
\label{sec:data_curation}
%++++++++++++++++++++++++++++++++++++++++++++++++

First, we collect a bucket of candidate e-commerce datasets. Most datasets are inherently designed for text-only or image-only tasks. 
To ensure the diversity of the modalities, we exclude those without product image information or rich text information. 
In addition, to collect real-world multimodal e-commerce data, we focus on the dataset derived from the Amazon platform, including Amazon Review~\cite{ni2019justifying, hou2024bridging}, AmazonQA~\cite{gupta2019amazonqa}, MAVE~\cite{yang2022mave}, and Shopping Query Dataset~\cite{reddy2022shopping}. This stage results in 5 raw datasets. 
Then we formulate the tasks in a multimodal way to ensure both text and image are meaningful for the tasks. The detailed curation process is in the Appendix.
\end{comment}
% \vspace{-1pt}
%
% \vspace{3pt}
%++++++++++++++++++++++++++++++++++++++++++++++++
\subsection{Quality Control}
\label{sec:dataset:quality}
%++++++++++++++++++++++++++++++++++++++++++++++++

% \xia{@Bo and Xinyi: need significant improvement on this section}

%In this section, we introduce our phased data cleaning to further control the quality of our data.
% In \dataset, we carry out the following procedures to
% ensure its accuracy and high quality, following the principle described in ECInstruct~\cite{peng2024ecellm}.
% \zz{confused by the two following and besides. Do you want to say: 
% In \dataset, to
% ensure its accuracy and high quality, we follow the principle described in \mbox{ECInstruct}~\cite{peng2024ecellm}.

In constructing \dataset, we adopt established principles from other instruction datasets~\cite{peng2024ecellm, fang2024molinstructions, yue2023mammoth}, focusing on clear instructions, consistent data formatting, and good alignment between input and target outputs~\cite{gadre2023datacomp}.
Those are critical for training generalizable instruction-following models.

% \xia{need to cite more and from other domains, principled work} \xia{need to briefly talk about what pricinples?}
% }
%
Besides, 
we exclude products without an accompanying image available to ensure all modalities are consistently available.
We select medium-size images
(500$\times$500 resolution) for each product to balance visual clarity and computational efficiency.
%Typically in the Amazon platform, each product is equipped with several images. 
%We utilize the cover image of 500$\times$500 resolution as the product image representation. 
%
We retain only products that include both detailed textual descriptions and corresponding images to ensure sufficient multimodal information for effective foundation model training.
% \xia{awkward langauge -- rephrase; unify style: detailed data vs images available}
% We also control both the product text information and within a reasonable length.
% keep texts and captions within a reasonable length;
%
% For example, in the \SR task, we truncate the long product text into 25 words; in the \SA task, we remove the product reviews that are less than 10 words. 
% For all the captions, we only retain the first 30 words. 
%
% In addition, we remove \hl{overlapping data} \xia{what do you mean by this?} between training and test sets to avoid data leakage, 
In addition, we remove samples from the test sets that also appear in the training set to prevent data leakage and ensure a clean separation for both IND and OOD evaluations.
% especially for OOD sets where product categories in training and test sets are completely disjoint.
% For instance, we only retain the first 50 interactions in the \SR task.
%
We further conduct manual scrutiny on the 1,000 randomly sampled instances~\cite{hedt2013effect} to ensure the overall data quality of accuracy, clarity, and relevance.
Only products with both high-quality images and detailed textual descriptions are retained to support effective multimodal learning.
%
% Overall, the data quality is assured by both carefully designed scripts and manual sampling inspection. 
This rigorous quality assurance process ensures that \dataset provides a reliable and standardized dataset for evaluating MFMs in e-commerce.
% \xy{More details of quality control can be found in eCeLLM~\cite{peng2024ecellm}.}
%
% the \hl{generated product captions are reasonable} \xia{what is this?} and the curated samples cover \hl{a diverse knowledge range} \xia{what is this?}.
% Due to the computing resource limitation
% Finally, we format all the instances with well-defined instructions.
% with 5K out-of-domain test instances and 70K in-domain instances, which split the train, validation, and test set with a ratio of 8:1:1.
Details of the dataset processing are in Appendix~\ref{sec:appendix:dataset}. 
%

% \hl{details in Appendix?}
% \vspace{-25pt}
%++++++++++++++++++++++++++++++++++++++++++++++++
\subsection{Dataset Partitioning} %\xia{need a better title}}
\label{sec:split}
%++++++++++++++++++++++++++++++++++++++++++++++++
% \vspace{-2pt}

% We split \dataset into the training set, validation set, in-domain (IND) test set, and out-of-domain (OOD) test set as 
% \xia{move this table earlier as table 1}.
% \xia{@Xinyi and Bo, I think we can simplify this section: IND and OOD, by refering to eCellM 
% paper, and move IND and OOD details to Appendix}

% We follow ECInstruct~\cite{peng2024ecellm} to split training sets, validation sets, in-domain (IND) test sets, and out-of-domain (OOD) test sets, detailed in Appendix~\ref{sec:appendix:data}. Total \dataset contains 75K samples and is summarized in Table~\ref{tbl:data_summary}.
Raw datasets of the \CC, \PRP, and \SA tasks are first split into training, validation, and test data at 8:1:1 ratio. 
For the \AP, \PSI, and \MPC tasks, the raw datasets are already split. 
For the \SR task, we follow the convention~\cite{hou2022towards}, leaving the last products in sequence interactions as the test data and the second last products as validation data.
Table~\ref{tbl:data_summary} summarizes the different splits. 

\noindent
\textbf{Training Set}~~
\dataset contains 8K samples for each individual task. These are combined into a single set of 56,000 samples, forming the complete training set for \dataset.

\noindent
\textbf{Validation Set}~~
\dataset includes a validation set of 1K samples for each individual task. These validation sets are combined into a single set of 7,000 samples, forming the complete validation set for \dataset.

\noindent
\textbf{In-domain (IND) Test Set}~~
For each of the 7 tasks, \dataset also includes an in-domain test set consisting of 1K samples. IND is defined in terms of products that belong to the same set of categories as those used in the training set.

\noindent
\textbf{Out-of-domain (OOD) Test Set}~~
To assess the generalizability of models to unseen samples and address the cold-start issue~\cite{schein2002methods,lika2014facing} in e-commerce, we create OOD test sets in \dataset. OOD is defined as new products that are not seen during training, identified by their category information. 
Five tasks (\AP, \CC, \PRP, \SA, and \SR) 
have products from various categories. Samples from certain categories are held out as OOD sets. 
% \xia{what does this mean exactly?} 
% Five tasks (\AP, \CC, \PRP, \SA, and \SR) are simulated with OOD scenarios by holding out entire product categories, i.e., during training, models do not see any products from these categories, and they are only introduced at test time. This setup evaluates the model's ability to generalize to entirely new product types not encountered during training.
% (with each product belonging to only one category).
We focus on new products instead of new users because user identifiers are anonymous in the dataset.

\begin{table}[!t]
  \centering
  \begin{footnotesize}
  \begin{threeparttable}
      \begin{tabular}{
	@{\hspace{0pt}}l@{\hspace{3pt}}
	@{\hspace{3pt}}c@{\hspace{3pt}}
	@{\hspace{3pt}}c@{\hspace{3pt}}
	@{\hspace{5pt}}c@{\hspace{5pt}}
	@{\hspace{5pt}}c@{\hspace{0pt}}
      }
      \toprule
       \textbf{Tasks} & \textbf{Training}  & \textbf{Validation}  & \textbf{IND}  & \textbf{OOD}  \\
	 \midrule
       %
       % Answerability Prediction (
       \AP, \CC, \PRP, \SA, \SR & 8,000 & 1,000 & 1,000 & 1,000  \\
%       % Category Classification (
%       \CC & 8,000 & 1,000 & 1,000 & 1,000  \\
%       % Product Relation Prediction (
%       \PRP & 8,000 & 1,000 & 1,000 & 1,000  \\
%       % Product Substitute Identification (
       \PSI, \MPC & 8,000 & 1,000 & 1,000 & \ding{55}\\
       % Query-product Relevance Classification (
%       \MPC & 8,000 & 1,000 & 1,000 & \ding{55} \\
%       % Sentiment Analysis (
%       \SA & 8,000 & 1,000 & 1,000 & 1,000  \\
%       % Sequential Recommendation (
%       \SR & 8,000 & 1,000 & 1,000 & 1,000  \\
       \midrule
       \dataset & 56,000 & 7,000 & 7,000 & 5,000\\
      \bottomrule
      \end{tabular}
      % \begin{tablenotes}[normal,flushleft]
      % \begin{footnotesize}
      % \item 
      % In this table, IND and OOD refer to the in-domain evaluation and out-of-domain evaluation, respectively. 
      % \par
      % \end{footnotesize}
      % \end{tablenotes}
  \end{threeparttable}
  \end{footnotesize}
  %\vspace{-3pt}
  \caption{Summary of the \dataset dataset. IND and OOD refer to the in-domain evaluation and out-of-domain evaluation, respectively. }
  \label{tbl:data_summary}
  %\vspace{-8pt}
\end{table}

\subsection{High-quality Instructions}
\label{sec:dataset:instruction}
%++++++++++++++++++++++++++++++++++++++++++++++++

High-quality instructions are critical to the effective adaptation of general-purpose LLMs to e-commerce~\cite{peng2024ecellm,jin2024amazonm2,jin2024shoppingmmlu}
% \xia{stop refering this citation ALL the times!!!}.
%
In \dataset, to ensure its high quality, we carefully craft a instruction for each of the seven e-commerce tasks.
Each instruction has been meticulously evaluated and refined by human experts to ensure clarity, conciseness, and accuracy.
The detailed description of instructions is in Appendix~\ref{sec:appendix:instr}.%We provide an instruction for each task in \dataset detailed in Appendix~\ref{sec:appendix:instr}.
% \xia{how about template diversity?}
% \label{{tbl:data_summary}}
%%%%%%%%%%%%%%%%%%%%%%%%%%%%%%%%

%%%%%%%%%%%%%%%%%%%%%%%%%%%%%%%%
%%%%%%%%%%%%%%%%%%%%%%%%%%%%%%%%%%%%%%%%%%%%%%%%%
\section[]{\pipeline: Lightweight Learning Framework for E-commerce MFMs}
%Unify Vision-Language Content/Modality/Information for e-Commerce 
\label{sec:method}

\dataset presents a multimodal dataset designed to evaluate how well models can effectively leverage both visual and textual information for e-commerce tasks.
% \xia{awkward opening -- revise}
%
While directly fine-tuning general multimodal models may seem like a straightforward solution, 
the results of the fine-tuned MFMs (discussed in Section~\ref{sec:ind}) indicate 
that these models struggle with domain-specific challenges.
% \hl{Existing multimodal models process images in a context-agnostic manner}~{\citep{li2024llava-next-interleave}},
% \hl{treating all images as equally informative, failing to distinguish the noise from helpful information within images,
% and leading to ineffective aligned multimodal representations for e-commerce.}\xia{this sentence should be moved earlier in intro to 
% motivate the work}
%
% Fine-tuning an existing MFM does not inherently solve these challenges.
%
To address this, we introduce \pipeline, 
%, a framework
%designed to overcome these limitations in e-commerce scenarios.
% by \hl{generating context-aware captions, 
% filtering out ineffectual visual data, 
% and strategically integrating multimodal information.}\xia{this should be mentioned and highlighted in Intro}
%
% \st{By incorporating textual information with task-relevant visual data, {\pipeline} enhances multimodal understanding and improves performance across diverse e-commerce tasks.}
%
which consists of three key modules:
\textbf{(1)} 
% an enriched \mbox{context-conditioned} captioning module that generates context-conditioned captions from images 
an enriched module (\ECCC) that generates context-conditioned captions from images
(Section~\ref{sec:method:captioning}), 
\textbf{(2)} 
% a caption quality evaluation 
% module that verifies caption qualities 
a light-weighted module (\CQE) that evaluates caption qualities (Section~\ref{sec:method:selection}), and 
\textbf{(3)} a light-weighted multimodal 
information fusion module that integrates high-quality captions with item context information 
(Section~\ref{sec:method:fusion}) to perform e-commerce tasks.
Figure~\ref{fig:caslie} presents an overview of \pipeline. 
We provide an analysis in Appendix~\ref{sec:appendix:analysis} to explore 
the impact of captioning models in \ECCC and caption quality evaluation models in \CQE 
on the performance of \pipeline.

\vspace{-3pt}
%++++++++++++++++++++++++++++++++++++++++++++++++
\subsection{\mbox{Enriched Context-conditioned Captioning}}
\label{sec:method:captioning}
%++++++++++++++++++++++++++++++++++++++++++++++++
% \vspace{-1pt}

\pipeline first employs a novel enriched context-conditioned captioning module -- \ECCC,
to generate textual captions for images, conditioned on the corresponding context,
such as user queries or reviews.
% , \hl{i.e., with respect to different user queries on the same product, the textual captions for the images 
% of the product can be different. } \xia{What does this mean? also, DO NOT use ``i.e.," in main text, 
% it is used only in parentheses.}
% the e-commerce task involving the item (and other related items), 
% etc. 
%\hl{An example of the context-conditioned captioning is shown in Figure}~\ref{xxx}.
%\xia{place holder; not sure if we need this in the end}
%
% employs a powerful MFM model (e.g., \mbox{Llama-3.2-11B-Vision-Instruct}) as the captioner to adaptively generate a textual representation (i.e., caption) for each given image conditioned on the context (e.g., user query). 
%
%An example of the context-conditioned captioning is shown in Figure~\ref{xxx}.
%as follows:
%\begin{equation}
%    \label{eqn:caption}
%    \mathcal{T}=\mathtt{Captioner}(\mathcal{C},\mathcal{I})
%\end{equation}
%
Unlike CLIP-based models~\cite{chia2022fashionclip, stevens2024bioclip}, 
which implicitly assume that the image in its entirty is relevant to the context.
%that treat entire images as aligned with textual information, 
%assuming the images
%are useful in their entirety
%and fully coordinated with text,
% which may not be particularly true in many e-commerce applications, 
%
\ECCC selectively highlights image details pertinent to the given context. 
\ECCC utilizes 
the strong image understanding capability of
% a powerful 
pre-trained MFMs 
for conditioned caption generation via zero-shot prompting, 
%
% \st{It integrates} \xia{
integrating context information with well-elaborated instructions to form a prompt (detailed in Appendix~\ref{sec:appendix:instr}).
A unique advantage of using pre-trained MFMs  
is that their extensive world knowledge,
allowing \ECCC to enrich captions with relevant insights beyond what is explicitly visible in the images, and thus, benefiting target tasks.
%
% . Therefore, the generated captions can be enriched 
% with such knowledge that may not be presented explicitly in the images but is relevant 
% to the image details and beneficial to the target task. 
%
%In addition, as shown in Figure~\ref{xxx}, by leveraging a powerful MFM as the captioner,
%\pipeline integrates extensive world knowledge from the MFM. 
%
%This could improve captions with additional information that is related to the objects or details in images and benefits the target task but is not explicitly conveyed by the image.
%
We use Llama-3.2-Vision-Instruct~\cite{dubey2024llama3} as the \ECCC model.
\subsection{\mbox{Caption Quality Evaluation}}
\label{sec:method:selection}
%++++++++++++++++++++++++++++++++++++++++++++++++

Existing multimodal e-commerce methods use all 
available images equally~\cite{zhuge2021kaleido, gao2020fashionbert} 
without evaluating their potential contributions to the target tasks. 
We denote this strategy as \allcap (\underline{u}se \underline{i}t \underline{a}lways).
However, 
%as shown in %Figure~\ref{XXX}\xia{not sure if we need the figure}, 
not all product images are high-quality or contain pertinent information, 
particularly under different contextual conditions.
%\hl{For example, an image showing a happy user may not convey the specifics of the items that users 
%care about most as expressed in their reviews. }
%\xy{For example, a user sharing a review image of a messy kitchen might simply be trying to convey that the purchased utensils are useful}
%
To ensure that the visual data contributes effectively and meaningfully in different conditions,
\pipeline incorporates a caption quality evaluation module -- \CQE, 
to assess whether the generated captions, and thus the corresponding product images,
% whether\xia{ and how} the generated captions\xia{, and thus the corresponding product images, }
% , as described by their generated captions, 
%are useful.
meaningfully contribute to the task and should be utilized.

%\bo{
%Existing multimodal e-commerce methods~\cite{xxx} generally leverage all available images for the target task. 
%%
%However, as shown in Figure~\ref{xxx}, images (e.g., product images or user-uploaded images) in e-commerce applications may not always provide helpful information for the target task.
%%
%Incorporating images that do not offer beneficial information can introduce noise and lead to sub-optimal performance in the target task, as demonstrated in the literature~\cite{xxx}.
%%
%To avoid including less informative images, \pipeline deliberately evaluates whether an image provides additional information relevant to the target task, 
%based on its textual representation (e.g., caption) and conditioned on the context (e.g., product title).}
%%

\CQE evaluates caption qualities by determining whether or not the captions provide 
beneficial information 
% within the given context 
for the target task via binary classification.
It employs powerful LLMs
% (e.g., Llama-3.1-8B-Instruct) or 
and MFMs
% (e.g., Llama-3.2-11B-Vision-Instruct) 
as classifiers,
leveraging the contextual information and well-curated instructions (detailed in Appendix~\ref{sec:appendix:instr}) for zero-shot evaluations, predicting if the generated 
caption should be utilized.
% \xia{need a summary of the methods (e.g., prompting, in-context learning) here 
% and refer to the details in the Appendix}. 
%
%
To mitigate inconsistencies in LLM-based predictions~\cite{bonagiri2024measuring}, 
\CQE aggregates outputs from five LLMs via \underline{m}ajority \underline{v}oting, denoted as \majvote, 
to reach a consensus as the final decision.
\pipeline integrates only captions deemed beneficial, 
% with other textual information, 
enabling a more strategic and deliberate fusion of multimodal data.
We use five generalist models as the binary classifiers for \majvote: 
Llama-3.2-3B-Instruct, 
Llama-3.1-8B-Instruct, 
and Llama-3.2-Vision-Instruct~\cite{dubey2024llama3}, as well as  
Mistral-7B-Instruct-v0.3~\cite{jiang2023mistral}, and 
Phi-3.5-mini-Instruct~\cite{abdin2024phi3}.

\subsection{\mbox{Modality-unified E-commerce Module}}
\label{sec:method:fusion}
%++++++++++++++++++++++++++++++++++++++++++++++++

% \xia{need to discuss this section!}

Through
\ECCC and \CQE, \pipeline explicitly translates visual content (i.e., images) into useful textual representations (i.e., captions).
These textual representations can be seamlessly integrated with other textual information (e.g., product titles or user reviews) 
by concatenating them.
Such concatenated texts will be used as input and the corresponding response as output to fine-tune a
modality-\underline{uni}fied e-co\underline{MM}erce \underline{M}odule, denoted as \uniMMM.
% \xia{it is unclear how integrate. need details}
%
%\xia{need a summary on how this is implemented, for example, through instruction-tuning or prompting, and refer to details in the Appendix.}
%
%Such a module is denoted as {\uniMMM} (\hanwen{modality-\textbf{uni}fied e-co\textbf{MM}erce \textbf{M}odule}). 
%\xy{We utilized the concatenated text as input and corresponding response as output to fine-tune {\uniMMM}.}
%
%Compared to existing MFMs that fuse visual and textual information by 
%\hl{embedding each modality, optimizing their alignment, and training customized fusion models, }
%\xia{double check; this might be a strong statement}
%%
%\uniMMM offers a simple, light-weighted, \mbox{(pre-)training-free} yet effective fusion framework, 
%enabling a unified view of multimodal data that can be easily used by any LLM-based
%e-commerce methods.
%
Three variants with various sizes for \uniMMM are fine-tuned:
\textbf{(1)} \methodL with Llama-2-13B-chat~\cite{touvron2023llama2}, 
\textbf{(2)} \methodM with Mistral-7B-Instruct-v0.3~\cite{jiang2023mistral}, and 
\textbf{(3)} \methodS  with Llama-3.2-3B-Instruct~\cite{dubey2024llama3} as the base models, 
respectively.
These models are optimized using LoRA~\cite{hu2022lora} and Huggingface transformers library~\cite{wolf2019huggingface} 
on the \dataset dataset.
%\hl{After applying {\ECCC} and {\CQE}}\xia{what does this mean???}, 
We refer to these models fine-tuned with the \pipeline learning framework as \pipelineL, \pipelineM, and \pipelineS, respectively.
\section{Experimental Setup}
\label{sec:exp_setup}
%%%%%%%%%%%%%%%%%%%%%%%%%%%%%%%%%%%%%%%%%%%%%%%%%

\paragraph{Baselines}
We compare \pipeline against 4 categories of baseline methods. 
\textbf{(1)} fine-tuned MFMs: LLaVA-Interleave~\cite{li2024llava-next-interleave}, 
\textbf{(2)} e-commerce LLMs: eCeLLM-L and eCeLLM-M~\cite{peng2024ecellm}, 
\textbf{(3)} fine-tuned CLIP-based models: FashionCLIP~\cite{chia2022fashionclip}, and
% \textbf{(2)} fine-tuned LLMs, 
\textbf{(4)} textual task-specific models. 
% To evaluate the \SR and \CC tasks,  
% we fine-tune \mbox{RECFORMER}~\cite{li2023recformer}, 
% and Sentence-BERT~\cite{reimers2019sbert}.
% All other tasks are evaluated on the 
% fine-tuned DeBERTa~\cite{he2021deberta}.
More detailed experimental setup is reported in Appendix~\ref{sec:appendix:exp_setup}.
We conduct IND and OOD evaluation (Section~\ref{sec:dataset}) for all the methods. 
The fine-tuned models and textual task-specific models are trained on \dataset.
More details on the experimental setup are available in Appendix~\ref{sec:appendix:exp_setup}.
% \xia{make it clear they are fine-tuned on our data.}

% \vspace{-2pt}

% \paragraph{\xy{CQE Settings}}
% % \xia{need a different title}}
% In \CQE, we use five \xy{generalist} models as the binary classifiers for \majvote: 
% \textbf{(1)} Llama-3.2-3B-Instruct~\cite{dubey2024llama3}, 
% \textbf{(2)} Llama-3.1-8B-Instruct~\cite{dubey2024llama3}, 
% \textbf{(3)} Llama-3.2-Vision-Instruct~\cite{dubey2024llama3}, 
% \textbf{(4)} Mistral-7B-Instruct-v0.3~\cite{jiang2023mistral}, and 
% \textbf{(5)} Phi-3.5-mini-Instruct~\cite{abdin2024phi3}.
% %
% \xia{are they fine tuned or not?}
%\xia{We do not fine-tune these models -- Xinyi, please finish/revise.}

% \subsection{Metrics}
% For the \AP and \PSI tasks, we use the F1 score as the primary metric. For the \CC and \SR tasks, we evaluate the results using Recall@1. The \MPC task is evaluated by accuracy. The \PRP and \SA are primarily evaluated by macro F1 score. The detailed results of each task are reported in Appendix~\ref{sec:appendix:full_results}. All the tasks undergo a systematic and rule-based evaluation pipeline.

% \subsection{Settings}
% We build \method series models on top of 3 base models: Llama-2-13B-chat~\cite{touvron2023llama2}, Mistral-7B-Instruct-v0.3~\cite{jiang2023mistral}, and Llama-3.2-3B-Instruct~\cite{dubey2024llama3}. 

%%%%%%%%%%%%%%%%%%%%%%%%%%%%%%%%%%%%%%%%%%%%%%%
\vspace{-1pt}
%%%%%%%%%%%%%%%%%%%%%%%%%%%%%%%%%%%%%%%%%%%%%%%
\section{Experimental Results}
\label{sec:exp_results}
%%%%%%%%%%%%%%%%%%%%%%%%%%%%%%%%%%%%%%%%%%%%%%%
\vspace{-2pt}
We conduct a systematic evaluation of \pipeline against all the baselines using the test set of each individual task in \dataset. 
% \hl{All experiments are conducted with NVIDIA A100 GPUs.} \xia{why mention it here? is it the most important thing that you have to mention early?}
%We evaluate the models using the test sets of individual tasks from \dataset. 
%
For a comprehensive evaluation, we utilize multiple metrics on each task.
To enable a succinct presentation, for each task, we present only the performance at the primary metric, defined as follows: 
F1 score for \AP and \PSI, Recall@1 for \CC and \SR, accuracy for \MPC, macro F1 score for \PRP and \SA. 
Complete results for each task are reported in Appendix~\ref{sec:appendix:full_results}.
When comparing {\pipeline} with baselines, 
we report the mean of {\pipeline}'s improvement over baselines per task as its overall improvement.
Additional results on the in-domain evaluation and complete evaluation results 
for all the e-commerce tasks are available in \ref{sec:appendix:full_results}.

\begin{table*}[htbp]
  \centering
  %\vspace{-10pt}
  \begin{footnotesize}
  \begin{threeparttable}
      \begin{tabular}{
        @{\hspace{0pt}}l@{\hspace{4pt}}
	  @{\hspace{4pt}}c@{\hspace{4pt}}%macrof1
	  @{\hspace{4pt}}c@{\hspace{4pt}}%f1
	  @{\hspace{4pt}}c@{\hspace{4pt}}%macrof1
	  @{\hspace{4pt}}c@{\hspace{4pt}}%hr1
	  @{\hspace{4pt}}c@{\hspace{4pt}}%f1
	  @{\hspace{4pt}}c@{\hspace{4pt}}%macrof1
	  @{\hspace{4pt}}c@{\hspace{0pt}}%f1
        @{\hspace{4pt}}c@{\hspace{4pt}}%None
	  @{\hspace{4pt}}c@{\hspace{4pt}}%macrof1
	  @{\hspace{4pt}}c@{\hspace{4pt}}%hr1
	  @{\hspace{4pt}}c@{\hspace{4pt}}%f1
	  @{\hspace{4pt}}c@{\hspace{4pt}}%macrof1
	  @{\hspace{4pt}}c@{\hspace{4pt}}%macrof1
      }
      \toprule
      \multicolumn{1}{c}{\multirow{2.5}{*}{\textbf{Model}}} 
      & \multicolumn{7}{c}{\textbf{IND}} & & \multicolumn{5}{c}{\textbf{
      OOD}} \\
      \cmidrule{2-8} \cmidrule{10-14}
      & \AP & \CC & \PRP & \PSI & \MPC & \SA & \SR & & \AP & \CC & \PRP & \SA & \SR \\
      % \cmidrule{2-8} \cmidrule{10-14}
      % & F1 & R@1 & M-F1 & F1 & Acc & M-F1 & R@1 & & F1 & R@1 & M-F1 & M-F1 & R@1\\
      \midrule
          
      \texttt{ft}-LLaVA-Interleave & 0.791 & \underline{0.964} & \underline{0.568} & \underline{0.340} & \underline{0.721} & 0.561 & 0.053
      && 0.579 & 0.043 & 0.334 & 0.206 & 0.000\\
      
      \cmidrule{2-8} \cmidrule{10-14}

      eCeLLM-L & 0.872 & 0.870 & 0.519 & 0.178 & 0.706 & 0.613 & \underline{0.188} 
      && \underline{0.860} & 0.916 & 0.531 & 0.584 & \underline{0.304} \\
      eCeLLM-M & 0.864 & 0.890 & 0.492 & 0.131 & 0.719 & \underline{0.632} & 0.182 
      && 0.841 & \underline{0.942} & \underline{0.564} & \underline{0.624} & 0.302 \\
      
      \cmidrule{2-8} \cmidrule{10-14}
      
      \texttt{ft}-FashionCLIP & 0.759 & 0.863 & 0.497 & 0.201 & 0.605 & 0.323 & 0.145 & & 0.600 & 0.903 & 0.453 & 0.376 & 0.087 \\
      
      \cmidrule{2-8} \cmidrule{10-14}
    
      Task-specific Model & \underline{0.868} & 0.671 & 0.531 & 0.316 & 0.702 & 0.495 & 0.163
      && 0.849 & 0.658 & 0.447 & 0.510 & 0.210 \\

      \midrule
      \pipelineL & 0.868 & 0.969 & 0.473 & 0.268 & 0.706 & 0.651 & 0.190
      && 0.840 & 0.968 & 0.531 & 0.607 & 0.297\\ 
      \pipelineM
      & \textbf{0.891} & \textbf{0.979} & \textbf{0.566} & \textbf{0.398} & \textbf{0.731} & \textbf{0.656} & \textbf{0.223} 
      && 0.855 & \textbf{0.977} & \textbf{0.585} & 0.625 & \textbf{0.330}\\
      
      \pipelineS & 0.871 & 0.963 & 0.504 & 0.336 & 0.707 & 0.601 & 0.196
      && \textbf{0.857} & 0.959 & 0.580 & \textbf{0.647} & 0.297\\
      \cmidrule{2-8} \cmidrule{10-14}

      imprv over best (\%; avg: 5.3) &  2.6 & 1.6 & -0.4 & 17.1 & 1.4 & 3.8 & 18.6 &&
      -0.3 & 3.7 & 3.7 & 3.7 & 8.6\\
      % ~~average imprv (\%; avg: 50.3) & 5.7 & 11.0 & 10.8 & 76.6 & 5.2 & 23.8 & 61.6 
      % && 12.2 & 280.5 & 23.2 & 37.6 & 54.9 \\

      % ~~caption used (\%; avg: 45.0) & 75.6 & 23.7 & 50.5 & 20.9 & 38.0 & 68.8 & 28.6 
      % && 68.2 & 24.6 & 43.2 & 71.6 & 26.3 \\
      caption used (\%; avg: 45.0) & 62.1 & 62.3 & 50.5 & 74.5 & 72.2 & 56.8 & 30.3 
      && 68.2 & 62.6 & 43.2 & 56.4 & 30.4 \\

      \bottomrule
      \end{tabular}
      % \begin{tablenotes}[normal, flushleft]
      % \begin{footnotesize}
      % \item
      % The best performance of \pipeline is in \textbf{bold} and of baselines is in \underline{underline}.
      % The ``imprv over best'' refers to the improvement of \pipeline over the best baselines;
      % % ``average imprv'' refers to the average improvement of \pipeline over each baselines;
      % ``caption used'' refers to the percentage of captions selected by \majvote.
      % \par
      % \end{footnotesize}
      % \end{tablenotes}
  \end{threeparttable}
  \end{footnotesize}
  %\vspace{-3pt}
  \caption{Overall performance comparison. 
  The best performance of \pipeline is in \textbf{bold} and of baselines is in \underline{underlined}.
      The ``imprv over best'' refers to the improvement of \pipeline over the best baselines;
      % ``average imprv'' refers to the average improvement of \pipeline over each baselines;
      ``caption used'' refers to the percentage of captions selected by \majvote. }
  \label{tbl:performance_ind}
  %\vspace{-8pt}
\end{table*}

%\label{tbl:performance_ind}
\vspace{-2pt}
%======================================================================
\subsection{In-domain Evaluation}
\label{sec:ind}
%======================================================================

The left part of Table~\ref{tbl:performance_ind} shows the overall performance in IND evaluation. 
% \hl{By default, as discussed in Section~{\ref{sec:method}}, 
% we use Llama-3.2-Vision-Instruct as the captioner and 
% the employ 5 evaluators  
% (Llama-3.2-3B-Instruct, Llama-3.1-8B-Instruct, 
% Llama-3.2-Vision-Instruct, Mistral-7B-Instruct-v0.3, 
% and Phi-3.5-mini-Instruct) as {\majvote} results.}
% \bo{@Xinyi why do we want to talk about this?}
% \xy{Dr. Ning suggests we reiterate the model settings for the readers}
%
%\bo{In Table~\ref{tbl:performance_ind}, \allcap denotes the \pipeline variant that uses all available image captions, while \majvote refers to a method that predicts and decides whether or not to include an image caption by majority voting (Section~\ref{sec:method:captioning}).}
% \xy{this paragraph added following the previous comments
% \\xia{need to better describe Table 1 setup: what is used for caption generation and selection, respectively. where are the full results and do you have discussions on the full results? }
% and was removed now}
% \bo{@Xinyi As in eCeLLM, we may want to discuss the training set difference between \pipeline and baselines here.}
% \xy{We already discussed this in Section~\ref{sec:exp_setup} Eexperimental Setup}

% \paragraph{\st{Overall Comparison:}} %Overall, 
%
\textbf{(1)} \pipelineM
% \bo{@Xinyi if we use \pipeline here how to differentiate the models with different sizes?} \xia{how about $\ECCC$+$\allcap$+$\methodL$, or $\pipeline_{\allcap, \methodL}$}
\textbf{\textit{substantially outperforms the baselines at 6.4\%
% \xia{need to update and be more clear what this number is about} 
on across 7 tasks}}
%
% \bo{@Xinyi please clarify here again how the improvement is calculated.}
%
(average of the improvement on each task) as shown in Table~\ref{tbl:performance_ind}.
% using \majvote (Section~\ref{sec:method:selection}) strategy.
% \xy{(may use \methodM to refer to using \methodM-\majvote by default, and \methodM-\allcap to refer to when using-all-caption/use-it-always strategy)}
% Particularly, \pipeline models achieve superior performance 
% with 45.8\% improvement over the FashionCLIP, 
% 6.5\% improvement over the best fine-tuned LLMs (\textit{ft}-mistral-7B-v0.3),
% % \bo{@Xinyi I think it should be LLMs. Need to unify across the paper.}
% 52.9\% improvement over \textit{ft}-LLaVA-NExT-interleave, 
% and 22.1\% over SoTA task-specific models.
These results demonstrate the remarkable effectiveness of \pipeline compared
with the fine-tuned CLIP-based model, fine-tuned LLMs, e-commerce LLMs,
fine-tuned MFMs, and the task-specific
models across the widely-performed e-commerce tasks.
% \bo{@Xinyi can we present an average improvement here as in the eCeLLm paper, and move the comparison with different types of baselines to the discussion below?}

%
% \bo{\st{The textual representation of images can be seamlessly combined 
% with other textual data, creating a unified multimodal view 
% that enhances \pipeline's performance.}}

% \pipeline comprehensively learns the text and captioned image information while the \hl{CLIP-based model only learns aligned data through contrastive learning} 
% \xia{is this correct? doesn't it also learn the text and image information? I don't think the key difference you argued is to the point. which claim about your method does this comparison supports?}. However, multimodal information is \hl{unaligned in most e-commerce scenarios} \xia{this is only your statement and it is a very strong statement. Why unaligned? What is the key and accurate insight here?}. The remarkable improvement suggests that \hl{by integrating the text and image information together, {\pipeline} can learn all rich information without alignment limitations.} \xia{this is a wrong conclusion and does not echo back to the innovation you claim in method}

% \paragraph{\st{Comparison with Fine-tuned LLMs and e-commerce LLMs}}
% \xia{the order here and in the tables --  Fine-tuned LLMs and e-commerce LLMs, should be 
% the same as in your baseline description. change the orders in tables}
% \xy{sure.}
% \xy{(image information do help; jointly learn from multiple tasks.)}

% \paragraph{\st{Comparison with Fine-tuned MFMs}}
%
\textbf{(2)}
\textbf{\textit{\pipelineM achieves a considerable 
% 10.7\% 
52.9\%
improvement over 
the fine-tuned MFM {ft}-LLaVA-NExT-Interleave}}, 
as demonstrated in Table~\ref{tbl:performance_ind}.
Notably, 
% \textit{ft}-LLaVA-NExT-Interleave struggles significantly with the task \SR 
% that requires processing multiple images, 
% while \pipeline achieves state-of-the-art performance
% (0.053 vs 0.223). 
the most significant performance gap occurs on the \SR task (0.223 vs. 0.053), which involves processing multiple images.
% \hl{The result substantiates the flexibility of {\pipeline} to effectively process images and capture rich pertinent visual information, hence improving performance on e-commerce tasks.}
% \xia{This is not accurate -- let's discuss}
%
ft-LLaVA-NExT-Interleave
directly encodes raw images alongside text in a fixed interleaved format, treating all visual content indistinguishably regardless of context.
% \xia{and treating visual content and textual content indistinguishably. --@Xinyi, double check if this is accurate.}
On the contrary, 
\pipeline uses visual content differentially via context-conditioned captioning, 
%{\pipeline} leverages context-conditioned captions as the vision representation, 
emphasizing task-related information from images. 
This process enables \pipeline to focus on the most informative image content while discarding irrelevant or noisy inputs, leading to significantly better performance, particularly on complex tasks like \SR.
% The result substantiates the flexibility of {\pipeline} to effectively process multiple images and capture rich pertinent visual information by conditioned captioning. However, MFMs
%%
%\zz{@Bo, what is the best word to replace pixels here? MFMs use a vision encoder + connector to integrate visual features with LLM}
%
% and 
%eliminates vision content that is not beneficial to the e-commerce tasks. 
% \hl{{\pipeline} also helps avoid potential misalignment issues in MFMs, 
% when images do not convey information concordant with texts.} \xia{where is this from based on Table 4?}
%
% \hl{Additionally, {\pipeline} enriches the textual representation 
% of images by incorporating world knowledge, further enhancing its performance 
% compared to MFMs. }\xia{where is this from based on Table 4?}
%seamlessly integrating it with other textual data 
%to create a unified multimodal view that boosts performance on e-commerce tasks.

% \paragraph{\st{Comparison with CLIP-based Models:}}
% \xy{(seamlessly learn combined visual and textual information; leverage world knowledge; focus on the informative part of the image.)}
\textbf{(3)}
\textbf{\textit{\pipeline exhibits superior performance over
e-commerce LLMs}}.
Specifically, 
\pipelineM outperforms 
eCeLLM-L by 25.2\% and eCeLLM-M by 37.1\%.
The results highlight the benefit of incorporating contextually relevant product image information into \pipeline, while eCeLLM models only utilize textual data.

We provide more analysis on IND evaluation compared to e-commerce LLMs and task-specific models in Appendix~\ref{sec:appendix:ind_result}, as well as the comparison with proprietary models and the error analysis in Appendix~\ref{sec:appendix:close} and \ref{sec:appendix:error}. 
In general, \pipelineM outperforms both ft-FashionCLIP and task-specific models by 45.8\% and 22.1\% gains, respectively.
Moreover, the mid-size \pipelineM offers the best performance, benefiting from its powerful base model.
\vspace{-1pt}
%======================================================
\subsection{Out-of-domain Evaluation}
\label{sec:ood}
%======================================================
\vspace{-1pt}
% \label{tbl:performance_ts}

\setcounter{table}{5}
\begin{table*}
  \centering

  \begin{footnotesize}
  \begin{threeparttable}
      \begin{tabular}{
        @{\hspace{4pt}}l@{\hspace{4pt}}
        @{\hspace{4pt}}l@{\hspace{4pt}}
	  @{\hspace{4pt}}c@{\hspace{4pt}}%macrof1
	  @{\hspace{4pt}}c@{\hspace{4pt}}%f1
	  @{\hspace{4pt}}c@{\hspace{4pt}}%macrof1
	  @{\hspace{4pt}}c@{\hspace{4pt}}%hr1
	  @{\hspace{4pt}}c@{\hspace{4pt}}%f1
	  @{\hspace{4pt}}c@{\hspace{4pt}}%macrof1
	  @{\hspace{4pt}}c@{\hspace{0pt}}%f1
        @{\hspace{6pt}}c@{\hspace{6pt}}%None
	  @{\hspace{4pt}}c@{\hspace{4pt}}%macrof1
	  @{\hspace{4pt}}c@{\hspace{4pt}}%hr1
	  @{\hspace{4pt}}c@{\hspace{4pt}}%f1
	  @{\hspace{4pt}}c@{\hspace{4pt}}%macrof1
	  @{\hspace{4pt}}c@{\hspace{4pt}}%macrof1
      }
      \toprule
      \multicolumn{2}{c}{\multirow{2.5}{*}{\textbf{Setting}}} 
      & \multicolumn{7}{c}{\textbf{IND}} & & \multicolumn{5}{c}{\textbf{OOD}} \\
      \cmidrule{3-9} \cmidrule{11-15}
      && \AP & \CC & \PRP & \PSI & \MPC & \SA & \SR & & \AP & \CC & \PRP & \SA & \SR \\
      % \cmidrule{3-9} \cmidrule{11-15}
      % && F1 & R@1 & M-F1 & F1 & Acc & M-F1 & R@1 & & F1 & R@1 & M-F1 & M-F1 & R@1\\
      \midrule
    
      \multirow{3}{*}{\uniMMM} 
      % &\texttt{ft}-Llama-2-13B
      & {-$\mathtt{L}$} & 0.866 & 0.969 & 0.468 & 0.235 & 0.700 & 0.628 & 0.184 && 0.831 & 0.959 & 0.523 & 0.595 & 0.285 \\
      % &\texttt{ft}-Mistral-7B-v0.3 
      & {-$\mathtt{M}$} & 0.876 & 0.971 & 0.533 & 0.312 & 0.725 & 0.617 & 0.218
      && 0.847 & 0.965 & 0.530 & \textbf{0.659} & 0.312\\
      % &\texttt{ft}-Llama-3.2-3B
      & {-$\mathtt{S}$} & 0.866 & 0.951 & 0.493 & 0.270 & 0.699 & 0.565 & 0.191
      && 0.838 & 0.962 & 0.511 & 0.614 & 0.305 \\

      \midrule

      \multirow{3}{*}{\allcap}
      & {-$\mathtt{L}$} 
      & 0.859 & 0.973 & 0.486 & 0.268 & 0.704 & 0.607 & 0.135
      && 0.840 & 0.968 & 0.533 & 0.606 & 0.236 \\

      & {-$\mathtt{M}$}
      & 0.885 & 0.976 & 0.535 & 0.352 & 0.722 & 0.642 & 0.207
      && \textbf{0.859} & 0.976 & 0.532 & 0.613 & 0.310 \\

      & {-$\mathtt{S}$}
      & 0.869 & 0.958 & 0.503 & 0.299 & 0.702 & 0.578 & 0.196 
      && 0.856 & 0.957 & 0.515 & 0.565 & 0.280\\
      
      \midrule
      \multirow{3}{*}{\pipeline}
      % &\pipelineL
      & {-$\mathtt{L}$} & 0.868 & 0.969 & 0.473 & 0.268 & 0.706 & 0.651 & 0.190
      && 0.840 & 0.968 & 0.531 & 0.607 & 0.297\\ 
      % &\pipelineM
      & {-$\mathtt{M}$} & \textbf{0.891} & \textbf{0.979} & \textbf{0.566} & \textbf{0.398} & \textbf{0.731} & \textbf{0.656} & \textbf{0.223} 
      && 0.855 & \textbf{0.977} & \textbf{0.585} & 0.625 & \textbf{0.330}\\
      
      % &\pipelineS 
      & {-$\mathtt{S}$} & 0.871 & 0.963 & 0.504 & 0.336 & 0.707 & 0.601 & 0.196
      && 0.857 & 0.959 & 0.580 & 0.647 & 0.297\\

      \bottomrule
      \end{tabular}
      % \begin{tablenotes}[normal, flushleft]
      % \begin{footnotesize}
      % \item
      % The best performance on each task is in \textbf{bold}.
      % \par
      % \end{footnotesize}
      % \end{tablenotes}

  \end{threeparttable}
  \end{footnotesize}
  \caption{Ablation study on different module settings of \pipeline of large (-$\mathtt{L}$), middle (-$\mathtt{M}$) or small (-$\mathtt{S}$) sizes. \uniMMM is the ablated version that uses text-only input.
   \pipeline-\allcap is the ablated version that always uses the visual information without quality evaluation.
  The best performance on each task is in \textbf{bold}.} 
    \label{tbl:ablation}

\end{table*}

% \label{tbl:ablation}

% \input{table/performance_ood}
%\label{tbl:performance_ood}

% \xia{need an overall performance discussion}

The right part of Table~\ref{tbl:performance_ind} 
presents the performance of \pipeline 
and baselines in OOD evaluation. 
Overall,
% we observe a similar trend to that in IND evaluation. 
% \hanwen{why IND suddenly appears here? in-domain? and where are the experiments for in-domain?} \xy{in-domain evaluation is discussed in Section~\ref{sec:ind}, here `similar trend' refers to the outperformance relationships between our \pipeline and baselines. Now this part removed.}
\textbf{\textit{\pipeline demonstrates strong generalizability to 
handle products in new categories,}}
with \pipelineM outperforming the best baselines by 3.9\% average improvement.
% \xy{leveraging context-aware captioning and multimodal information, 
% enabling it to effectively address real-world e-commerce tasks.}
% \bo{@Xinyi I do not follow here}
% \hl{through leveraging comprehensive knowledge 
% and effectively handling real-world e-commerce tasks} 
% \xia{@Xinyi, what do you mean?}
% \xia{is this right? need more accurate languages}.
% \bo{@Xinyi Need to discuss the out-of-domain evaluation also in Section~\ref{sec:dataset}.}
% \xy{Done.}

%\st{We find that fine-tuned MFMs fall significantly 
%behind \pipeline \xia{(e.g., give an example here)}.}
%\st{underscoring the strong generalizability of \pipeline.}
%\st{Specifically, }
\textbf{(1)}
\textbf{\textit{\pipelineM surpasses the \textit{ft}-LLaVA-NExT-Interleave 
by a substantial 624.6\% improvement across 4 tasks except for \SR in the {OOD} setting}},
% \bo{@Xinyi the number does not look right to me. Do not think you can calculate improvement on all the 5 tasks as the performance on \SR is 0...}
underscoring the strong generalizability of \pipeline.
% 
% \hl{This demonstrates the limited ability of 
% fine-tuned MFMs to transfer 
% knowledge from the in-domain to out-of-domain scenarios. }
% \hl{
\textit{ft}-LLaVA-NExT-Interleave appears to be struggling to transfer knowledge effectively 
%from IND to 
in OOD scenarios, possibly due to that products from new categories may have very different images.
% or similar images but very different textual information.
%
\pipeline takes advantage of the well-known generalizability of LLMs~\cite{touvron2023llama2, jiang2023mistral, dubey2024llama3} to understand such new images by translating images to context-conditioned textual representations, and thus generalizes well. % to OOD scenarios.
% } \xia{check if this is accurate}

%Their reliance on pixel-based image representations 
%may limit their adaptability when faced with unseen or novel data, 
%as they may not capture the context-specific details necessary for broader applicability.
%
% \xia{just claim why MFMs is worse in generalizability, instead of ``demonstrates."}
%The results suggest that textual representation 
%could be more effective than pixels for enabling strong generalizability

% \bo{@Xinyi textual representation could be more effective than pixels for enabling strong generalizability. This could be an interesting observation and also explain why pre-training on images does not work as well as that in text.}

\textbf{(2)}
Similarly, \textbf{\textit{\pipelineM demonstrates significant advantages 
over \textit{ft}-FashionCLIP and 
% eCeLLM-M 
eCeLLM-L
in the OOD evaluation}}, 
with average 85.1\% and 6.4\% improvements, 
% with average improvements of \hl{45.8}\% and \hl{39.7}\%
% \xia{@Xinyi, double check!!!}, 
respectively.
\pipeline could easily leverage LLMs' generalizability and world knowledge that \textit{ft}-FashionCLIP doesn't enjoy. 
%Meanwhile, multimodal dataset \dataset enriches \pipeline's knowledge 
%that could be transferred to new news and new products.
%
Meanwhile, the ability to integrate visual information 
via context-conditioned captions allows \pipeline to better capture
% nuanced 
product details, enabling it to
understand new products more effectively
than \mbox{eCeLLM-M}, which focuses primarily on text-based information.
% \xia{@Xinyi and Bo, why is \pipeline better than eCellM in generalizability? need rationale}

% \xia{need to discuss}

%
% These suggest that \pipeline benefits from 
% \hl{both the high-quality multimodal e-commerce dataset {\dataset}} -- \xia{I don't get this point...} 
% the inherent generalizability of LLMs and the high-quality multimodal e-commerce dataset {\dataset}. 
\begin{comment}
As the cold-start issues remain a critical 
challenge in the e-commerce realm, 
the strong OOD generalizability 
of \pipeline to new scenarios, 
as shown in Table~\ref{tbl:performance_ind}, 
highlights its potential high utility 
for e-commerce applications.
\end{comment}
% exhibits potent transferability
% \xia{need to discuss this small section}

% \input{table/performance_generalist_vs_task-specific}
% % \label{tbl:performance_ts}
\vspace{-1pt}
%======================================================
\subsection[]{Task-Specific and Generalist \pipeline}
\label{sec:generalist}
%======================================================
\vspace{-1pt}
 \setcounter{table}{4}
\begin{table}
  \centering
  % \vspace{-9pt}
  \footnotesize
  \begin{threeparttable}
      \begin{tabular}{
        @{\hspace{0pt}}c@{\hspace{2pt}}
	  @{\hspace{2pt}}c@{\hspace{2pt}}
	  @{\hspace{2pt}}c@{\hspace{2pt}}%macrof1
	  @{\hspace{2pt}}c@{\hspace{2pt}}%f1
	  @{\hspace{2pt}}c@{\hspace{2pt}}%macrof1
	  @{\hspace{2pt}}c@{\hspace{2pt}}%hr1
        @{\hspace{2pt}}c@{\hspace{2pt}}%acc
        @{\hspace{2pt}}c@{\hspace{2pt}}%macrof1
        @{\hspace{2pt}}c@{\hspace{0pt}}%ndcg
      }
      \toprule
      % \multicolumn{3}{c}{\multirow{2}{*}{Model}} 
      \multirow{1}{*}{\textbf{Size}} 
      % & \multirow{2}{*}{Use Captions}
      & \multirow{1}{*}{\textbf{Training}} 
      & \AP & \CC & \PRP & \PSI & \MPC & \SA & \SR \\ % & \multirow{2}{*}{impro(\%)}\\
      % \cmidrule{3-9}
      % && F1 & R@1 & M-F1 & F1 & Acc & M-F1 & R@1 \\
      \midrule

      \multirow{2}{*}{-$\mathtt{L}$}
      & {T-spec.} & 0.837 & 0.931 & 0.428 & 0.000 & 0.671 & 0.553 & 0.058 \\
      & {Gen.} & 0.868 & 0.969 & 0.473 & 0.205 & 0.706 & 0.651 & 0.190 \\ 

      \midrule

      \multirow{2}{*}{-$\mathtt{M}$}
      & T-spec. & 0.866 & 0.968 & 0.495 & 0.000 & 0.709 & 0.600 & 0.197 \\
      & Gen. & 0.891 & 0.979 & 0.566 & 0.398 & 0.731 & 0.656 & 0.223 \\

      \midrule

      \multirow{2}{*}{-$\mathtt{S}$}
      & T-spec. & 0.838 & 0.912 & 0.460 & 0.000 & 0.684 & 0.557 & 0.121 \\
      & Gen. & 0.871 & 0.963 & 0.504 & 0.336 & 0.707 & 0.601 & 0.196 \\ 

      \bottomrule
      \end{tabular}
      % \begin{tablenotes}[normal, flushleft]
      % \begin{footnotesize}
      % \item
      % ``Task-spec.''/``Generalist''
      %  indicates the \pipeline models tuned on individual tasks/using all tasks together;
      % The best performance is in \textbf{bold}.
      % %
      % All {\pipeline} models use the \majvote strategy.
      % \par
      % \end{footnotesize}
      % \end{tablenotes}
  \end{threeparttable}
  \caption{Comparison of task-specific (T-spec.) and generalist (Gen.) \pipeline models of large (-$\mathtt{L}$), middle (-$\mathtt{M}$) or small (-$\mathtt{S}$) sizes.}  
  \label{tbl:performance_ts}
\end{table}

Table~\ref{tbl:performance_ts} presents the results of \pipeline fine-tuned with different strategies.
When comparing 
the task-specific \pipeline, which is fine-tuned for each individual task,
with the generalist \pipeline, which is fine-tuned across all the tasks together,
% under IND evaluation, 
% \xia{ordering in the table vs narratives!}
%
%When comparing \pipeline, which fine-tuned the \uniMMM models across all tasks (generalist), 
%with task-specific \pipeline which fine-tuned \uniMMM models on individual tasks under in-domain evaluation, 
we observe a trend consistent with that in prior research~\cite{peng2024ecellm}: 
\textbf{\textit{the generalist \pipeline outperforms task-specific \pipeline on each individual task}}. 
Generalist \pipelineL, \pipelineM, and \pipelineS exhibit significant improvements of 44.8\%, 7.3\%, and 15.4\% over 
their respective task-specific \pipeline across all tasks 
except for \PSI.
%\xia{need to explain why 0 for \PSI in task-specific setting}.
% \xy{Without knowledge transfer, LLMs only acquire limited information to conduct \PSI.}
% which is too hard for the task-specific {\pipeline} to acquire informative knowledge.
% \xia{need to be concrete here. 
% if in eCellM paper we already discussed the reason, can refer to it ``due to similar reasons as 
% discussed in XXX paper (section YYY)"}
These results highlight that
% the \hl{versatility of the {\pipeline} models, facilitating knowledge transfer across tasks, and significantly enhancing performance, particularly on complex e-commerce tasks like {\PSI}}.
% \bo{@Xinyi do not think versatility can facilitate knowledge transfer... Revise based on the eCeLLM paper. The point is}
training on all tasks together, \pipeline enjoys strong versatility and learns transferable knowledge across tasks to boost the performance on individual tasks.
% \xy{This improvement can be attributed to the transferable knowledge 
% learned during multi-task training. These captions facilitate knowledge transfer across tasks, allowing the generalist model to generalize better and boost performance on individual tasks }
%
It is noteworthy that on \PSI, all task-specific \pipeline models fail due to highly unbalanced labels (74\% negatives), whereas generalist \pipeline models still achieve considerable performance. 
This demonstrates that certain e-commerce tasks (e.g., \PSI) could substantially benefit from knowledge transfer through generalist modeling.
%

%\input{table/performance_transformability}
%% \label{tbl:performance_trans}

%%=========================================================
%\subsection{\xia{Multimodal Data vs Text-only Data for e-Commerce}}
%\label{sec:mul}
%%=========================================================
%
%\xia{need to discuss the conclusions here in this section!}
%
%Furthermore, as unveiled in Table~\ref{tbl:performance_trans},
%when comparing the performance between generalist models, 
%we find that \textbf{\textit{incorporating both vision and language information leads to 
%greater improvements than using text-only data for generalist models}}. 
%%
%Specifically, \textit{ft}-Mistral-7B-v0.3, %-- the base model of \pipeline's \uniMMM module, 
%which relies solely on language information, shows a 2.9\% average improvement over task-specific \textit{ft}-Mistral-7B-v0.3.
%% \bo{@Xinyi over what?}. 
%In contrast, \pipelineM that leverages both vision and language information achieves a 6.0\% average improvement over its task-specific version. 
%\st{This suggests that adding vision information through {\pipeline} significantly enhances the model's ability to transfer knowledge across diverse tasks.}
%\xy{By involving context-conditioned captions, 
%\pipeline can leverage broader world knowledge 
%to enhance visual information understanding. 
%The integration of visual and textual information leads to 
%a deeper and more robust knowledge transfer across diverse tasks, 
%suggesting that vision data enhances the model's versatility and ability to generalize beyond language alone.}
%% \xia{@Bo, need a better insight}

\vspace{-1pt}
%======================================================
\subsection[]{Ablation Study}
\label{sec:ablation}
%======================================================
\vspace{-1pt}
% \input{table/ablation}
% % \label{tbl:ablation}

% Due to the deterministic sequence of modules in \pipeline, 
In Table~\ref{tbl:ablation}, 
we compare the \pipeline framework with two ablated versions with selected modules: 
\uniMMM uses text-only input, and \pipeline-\allcap always uses the visual information without quality evaluation.
%
% with all modules (\ECCC-\CQE-\uniMMM, also denoted as \pipeline \xia{Very awkward}) and 
% with selected modules (\ECCC-\uniMMM, denoted as \pipeline-\allcap, and \uniMMM \xia{Very awkward}) 
% across various model sizes 
% (-$\mathtt{S}$, -$\mathtt{M}$, and -$\mathtt{L}$). 
%
Take the mid-size models as examples, 
% \ECCC-\uniMMM-$\mathtt{M}$ brings a 0.7\% average improvement compared to \uniMMM-$\mathtt{M}$, 
\textbf{\textit{\pipelineM brings 
a 4.1\% average improvement compared to \pipeline-\allcap-$\mathtt{M}$}}, 
and 
a 4.9\% average improvement over \uniMMM-$\mathtt{M}$, 
highlighting the importance of conditioned captioning and selective visuals integration. 
% \xia{Let's discuss the presentation}

% \xia{this paragrah needs rewritting. let's discuss}
These observations underscore 
the key benefits of \pipeline’s modular design
% to represent images through captioning (by \ECCC) and select beneficial visual information.
to integrate selective (by \CQE) text-based image representation (by \ECCC) into \uniMMM.
\pipeline gains benefit from \ECCC by extracting context-conditioned captions, effectively translating visual information into textual format for later seamless incorporation. 
The \CQE module further refines this process by filtering out non-beneficial image information, ensuring that only task-relevant visual data is integrated. 
By concatenating textual and selected visual information and feeding them into powerful \uniMMM,
% \xia{grammar}
{\pipeline} enhances its ability to jointly learn e-commerce tasks 
from a multimodal perspective, 
enabling performance that text-only information cannot achieve.

Besides, we also conduct ablation studies on using various captioning models in \ECCC and various evaluation strategies in \CQE, 
% detailed
demonstrating the effectiveness of our design 
in Appendix~\ref{sec:appendix:analysis}.
% The result shows that context-conditioned captioning models consistently yield higher improvements compared to standard, context-free captioning, suggesting that adaptively underscoring conditioned image details significantly enhances the model's performance.

% \bo{@Xinyi directly compare \pipeline with Llama-3.2-Vision-Instruct (context-conditioned caption) over that with Blip2.}
% \vspace{-2pt}
%%%%%%%%%%%%%%%%%%%%%%%%%%%%%%%%%%%%%
\section{Conclusion}
\label{sec:conclusion}
%%%%%%%%%%%%%%%%%%%%%%%%%%%%%%%%%%%%%
% \vspace{-2pt}

We develop and open-source a high-quality, multimodal instruction dataset for e-commerce.
%for developing and evaluating generalist MFMs in e-commerce. 
%
To our knowledge, \dataset is the first of its kind.
We also develop \pipeline, a simple, yet effective framework integrating multimodal information for e-commerce.
Leveraging \dataset, we fine-tune  
the state-of-the-art MFMs (\pipeline series) within \pipeline for e-commerce.
%
%a context-conditioned, light-weight \pipeline framework to integrate visual and textual information. The fine-tuned \uniMMM from the \pipeline framework can outperform the most advanced baseline models including
%
Our extensive evaluation of \pipeline models against the most advanced baseline models 
%including fine-tuned MFMs, and SoTA task-specific models,
%in both IND and OOD evaluations 
shows that \dataset enhances \pipeline with with advanced capabilities and strong generalizability in e-commerce applications.
%
% By providing a standardized benchmark and demonstrating its efficacy through \pipeline, we aim to facilitate advancements in the development of foundation models tailored for multimodal e-commerce applications.
% \texttt{ft}-Mistral-7B-v0.3 
% eCeLLM-M in both IND and OOD evaluations with improvements of 6.5\%, and 3.3\%, respectively.
%
%with an average improvement of 6.5\% in IND evaluation.
%
%\pipeline also achieves a notable average improvement of 3.3\% over the best baseline model ft-Mistral-7B-v0.3, highlighting its strong generalizability to OOD settings. 

% \clearpage
%%%%%%%%%%%%%%%%%%%%%%%%%%%%%%%%%%%%%
\section{Limitations}
%%%%%%%%%%%%%%%%%%%%%%%%%%%%%%%%%%%%%
%
First, while our dataset \dataset undergoes rigorous quality control, there remains a possibility that some samples may still contain noisy or inaccurate information (e.g., mismatch between text and image). This might hinder the performance of the \pipeline that is fine-tuned on this dataset. 
Second, the LLM-based captioning module \ECCC might generate inaccurate or even hallucinated captions in rare occasions, where the captions do not truthfully represent actual objects in
the images. This issue might be partially addressed via preference alignment and optimization~\cite{gunjal2024detecting} to reduce hallucination. 
Third, \CQE can only decide whether or not the captions provide beneficial information within the given context but lacks interpretability to explicitly pinpoint the particular regions/details of the images that are 
beneficial to the tasks. For future work, we plan to leverage image segmentation techniques~\cite{kirillov2023segment} to achieve a more fine-grained evaluation of the images. 
Fourth, our framework is based on manually-crafted prompt templates, which may be suboptimal in certain cases. For future work, we plan to introduce automatic prompt optimization techniques~\cite{pryzant-etal-2023-automatic} to create customized prompts tailored to various e-commerce tasks and use cases.

While it is our aspiration that e-commerce models can enrich users' online experience and enhance users' satisfaction, we also acknowledge that unintended use of e-commerce models might introduce popularity bias~\cite{chen2023bias} (e.g., only recommend popular products in the sequential recommendation task) among a large group of users. 
This issue might be exacerbated when the popular products have more, high-quality image
data, and thus bias the image data integration in multimodal e-commerce models. 
This issue can mitigated by introducing debiasing algorithms~\cite{wang2021deconfounded,zhang2021causal} in the future.

%%%%%%%%%%%%%%%%%%%%%%%%%%%%%%%%%%%%%
\section{Ethics Statement}
%%%%%%%%%%%%%%%%%%%%%%%%%%%%%%%%%%%%%

Our dataset \dataset is constructed all based on public, open-sourced datasets with proper licensing to allow for redistribution and research purposes (Table~\ref{tbl:license}). All the user IDs are fully anonymized, and there is no user profile information (e.g., user names, user address) that could lead to potential disclosure of user privacy.
%\section*{Acknowledgments}

\bibliography{paper}

\clearpage

\appendix
\setcounter{table}{0}
\setcounter{figure}{0}
\renewcommand{\thetable}{A\arabic{table}}
\renewcommand{\thefigure}{A\arabic{figure}}

%%%%%%%%%%%%%%%%%%%%%%%%%%%%%%%%%%%%%%%%%%%%%%%%%%%%%%%%%%
\section{Dataset Details}
\label{sec:appendix:dataset}
%%%%%%%%%%%%%%%%%%%%%%%%%%%%%%%%%%%%%%%%%%%%%%%%%%%%%%%%%%

\begin{table*}
  % \vspace{2pt}
  \centering
  \footnotesize
  \begin{threeparttable}
      \begin{tabular}{
        @{\hspace{2pt}}l@{\hspace{3pt}}
        @{\hspace{3pt}}l@{\hspace{3pt}}
        @{\hspace{3pt}}l@{\hspace{2pt}}
      }
      \toprule
      \textbf{Dataset} & \textbf{License Type} & \textbf{Source} \\
      \midrule
      Amazon Review & Not Specified & \href{https://amazon-reviews-2023.github.io/}{https://https://amazon-reviews-2023.github.io/} \\
      AmazonQA & Not Specified & \href{https://github.com/amazonqa/amazonqa}{https://github.com/amazonqa/amazonqa} \\
      MAVE & CC-by-4.0 & \href{{https://github.com/google-research-datasets/MAVE}}{https://github.com/google-research-datasets/MAVE} \\
      Shopping Queries Dataset & Apache License 2.0 & \href{https://github.com/amazon-science/esci-data}{https://github.com/amazon-science/esci-data} \\
      
      \bottomrule
      \end{tabular}
  \end{threeparttable}
  \caption{Details of Data Source License}
  \label{tbl:license}
\end{table*}

% \label{tbl:license}
To pursue adherence to data usage requirements, we check the licenses of \dataset data sources, ensuring their permission to publish. Table~\ref{tbl:license} presents the licenses of our curated dataset sources.

\subsection{Task Selection}
Following ECInstruct~\cite{peng2024ecellm}, \dataset comprises 7 widely-performed real-world tasks constructed from real-world data, which are ubiquitous and essential in the e-commerce domain elaborating in Table~\ref{tbl:tasks}.
Not all ECInstruct tasks are involved since some data sources lack vision information.
Previous methods for summarization~\cite{xu2020self, li2020keywords, li2020aspect}, extraction~\cite{zhu2020multimodal} and description generation~\cite{li2024multimodal} also aim for generation tasks in e-commerce domain but study different direction from this work. Therefore these tasks are not considered here.
Following prior research~\cite{wei2022finetuned} and taking into account the high computational demands, we uniformly downsample the training sets for each individual task to 8K samples, the validation sets to 1K, and the test sets to 1K. This ensures an optimal balance between data volume and efficient processing for affordable LLM evaluation.

\subsection{Data Selection}
In the \AP, \PRP, \SA, and \SR tasks, Tools category data from Amazon datasets~\cite{gupta2019amazonqa, hou2024bridging, ni2019justifying} serve as in-domain (IND) data sources, and Sports category data serves as out-of-domain (OOD) data.

For the \MPC and \PSI tasks, we directly process the row datasets~\cite{reddy2022shopping} from their original splits.

For the \CC tasks, we select the 100 most frequent fine-grained categories as in-domain (IND) data, while categories ranked between 100 and 200 in frequency are used as out-of-domain (OOD) data.

\subsection{Data Statistics}
\begin{figure*}
    \centering
    \subfloat[\AP]{\includegraphics[width=0.136\textwidth]{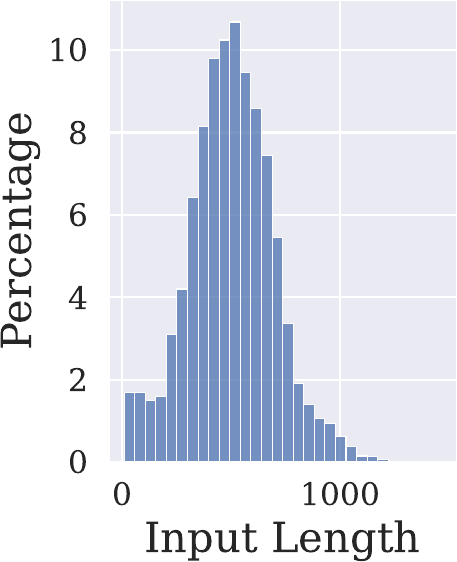}} \hfill
    \subfloat[\CC]{\includegraphics[width=0.136\textwidth]{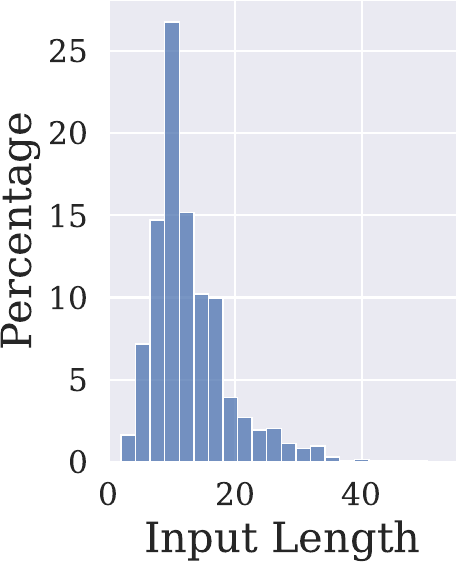}} \hfill
    \subfloat[\PRP]{\includegraphics[width=0.136\textwidth]{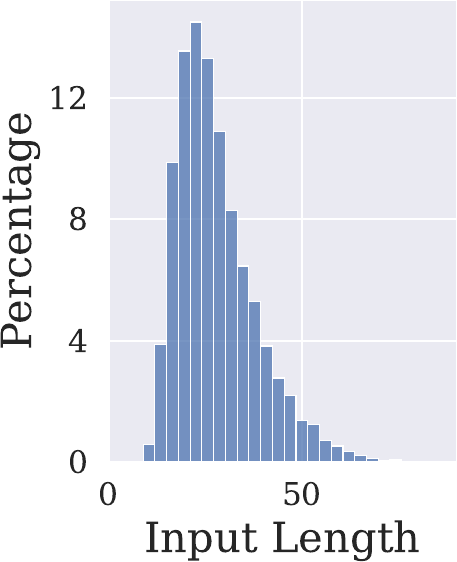}} \hfill
    \subfloat[\PSI]{\includegraphics[width=0.136\textwidth]{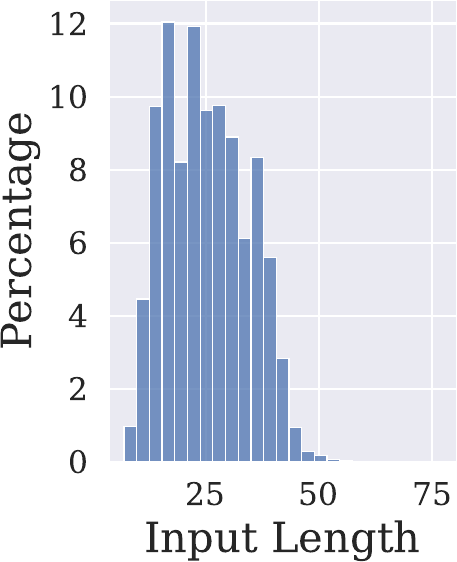}} \hfill
    \subfloat[\MPC]{\includegraphics[width=0.136\textwidth]{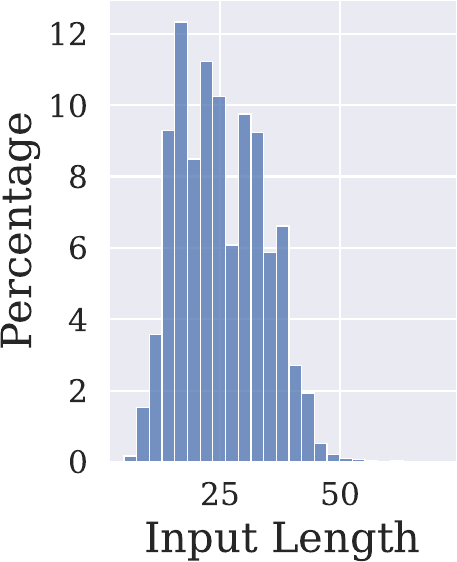}} \hfill
    \subfloat[\SA]{\includegraphics[width=0.136\textwidth]{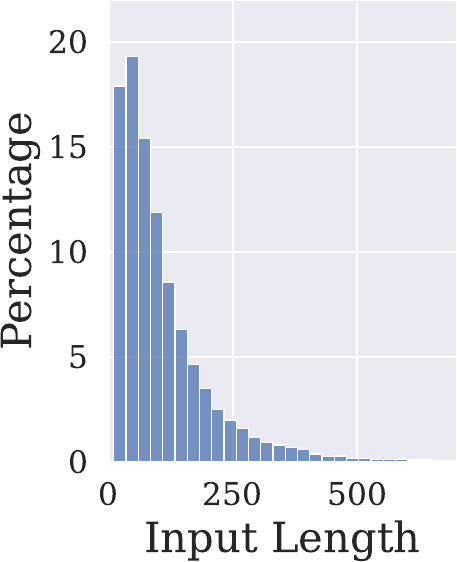}} \hfill
    \subfloat[\SR]{\includegraphics[width=0.136\textwidth]{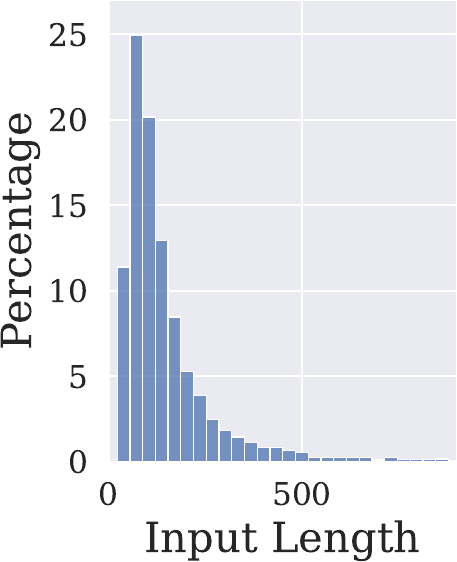}} 
    \caption{Distribution of Input Length in \dataset}
    \label{fig:7subfigures}
\end{figure*}

Figure \ref{fig:7subfigures} presents the distributions of input lengths for each task, measured by word count. For better clarity, we exclude very long inputs (those representing at most 1\% of samples) in the \SA and \SR tasks.

Table~\ref{tbl:category} presents the distribution of product categories in the \dataset dataset. The dataset spans a wide variety of categories, reflecting the heterogeneity of real-world e-commerce platforms. Notably, it includes high-volume categories and also incorporates lower-frequency and long-tail categories, enhancing its diversity. This stratified coverage across both popular and niche domains enables \dataset to support robust training and evaluation of multimodal models under varied product scenarios.

\begin{table}[ht]
\centering
\small
\begin{tabular}{l r}
\toprule
\textbf{Category} & \textbf{Percentage (\%)} \\
\midrule
Cell Phones and Accessories & 18.91 \\
Tools and Home Improvement & 14.82 \\
Electronics & 11.96 \\
Home and Kitchen & 11.02 \\
Clothing Shoes and Jewelry & 6.00 \\
Sports and Outdoors & 5.81 \\
Toys and Games & 4.65 \\
Books & 3.64 \\
Automotive & 2.83 \\
Beauty and Personal Care & 2.81 \\
Grocery and Gourmet Food & 2.75 \\
Health and Household & 2.68 \\
Patio Lawn and Garden & 2.06 \\
Office Products & 1.93 \\
Arts Crafts and Sewing & 1.48 \\
Pet Supplies & 1.47 \\
Others & 5.18 \\
\bottomrule
\end{tabular}
\caption{Category Statistics}
\label{tbl:category}
\end{table}

% \caption{Category Statistics}
% \label{tbl:category}

%%%%%%%%%%%%%%%%%%%%%%%%%%%%%%%%%%%%%%%%%%%%%%%%%%%%%%%%%%
\subsection{Data Processing}
% \label{sec:appendix:data}
%%%%%%%%%%%%%%%%%%%%%%%%%%%%%%%%%%%%%%%%%%%%%%%%%%%%%%%%%%
%
We conduct the data processing following \mbox{ECInstruct}~\cite{peng2024ecellm} as below. Besides that, we thoroughly check the availability of each product's image.
\subsection{Dataset Partitioning}
\paragraph*{Answerablity Prediction (\AP)}
\label{sec:appendix:ap}
We utilize the data from the Tools category of AmazonQA~\cite{gupta2019amazonqa} as the in-domain (IND) source and the Sports category as the out-of-domain (OOD) source for this task. The \textit{is\_answerable} annotations serve as the ground truth. In the structured dataset, the ratio of positive to negative samples is approximately 3:5.

\paragraph*{Category Classification (\CC)}
\label{sec:appendix:cc}
We use the fine-grained product category labels from MAVE~\cite{yang2022mave} as the ground truth. To ensure each selected category has sufficient data, we first sort the categories by frequency. We then select the 100 most frequent fine-grained categories as IND data, while categories ranked between 100 and 200 in frequency are designated as OOD data. Then we split IND data with an 8:1:1 ratio to formulate training, validation, and IND test set.

\paragraph*{Product Relation Prediction (\PRP)}
\label{sec:appendix:prp}
Similar to ECInstruct~\cite{peng2024ecellm}, to study product relationships, we utilize the product metadata from the Tools category as IND sources, with the Sports category serving as the OOD source. We collect product IDs from the metadata, removing any products that lack detailed information. Product titles and images are used to represent the products in this task, and any product pairs that appear multiple times with different relations are eliminated. After filtering and integrating the data with instruction templates, the three types of relationships (\textit{also buy, also view, and similar}) are distributed in the final dataset at approximately a 4:3:1 ratio.

\paragraph*{Product Substitute Identification (\PSI)}
\label{sec:appendix:psi}

We represent products from the Shopping Queries dataset~\cite{reddy2022shopping} using their titles and images and eliminate non-English samples. Each query-product pair is labeled into 4 categories (\emph{Exact, Substitute, Complement, and Irrelevant})
The query-product pairs with {\emph{Exact, Complement, or Irrelevant}} labels are relabeled as non-substitute.
The ratio of the positive and negative labels in the \dataset dataset is approximately 1:3.

\paragraph*{Multi-class Product Classification (\MPC)}
\label{sec:appendix:mpc}
The preprocessing of the \MPC is similar to that of \PSI, except that the \MPC is a multi-class classification task. The ratio of the four labels in the structured dataset (\emph{Exact, Substitute, Complement, and Irrelevant}) is approximately 20:7:1:4.

\paragraph*{Sentiment Analysis (\SA)}
\label{sec:appendix:sa}
For the sentiment analysis, we use the review data of the Tools category from the Amazon Review dataset~\cite{hou2024bridging} as the IND sources and the Sports category as the OOD source. We only retain the reviews that are longer than 10 words.

\paragraph*{Sequential Recommendation (\SR)}
\label{sec:appendix:sr}
In the \SR task, we utilize both product reviews and metadata from the Amazon Review dataset~\cite{hou2024bridging}. Additionally, we incorporate users' review histories as a representation of their interactions with products. The processing protocol follows the same steps as ECInstruct~\cite{peng2024ecellm}, with the primary distinction being the inclusion of images for each product. The curated dataset has an average of 10.7 interactions per user and an average text length of 18 words per product.
%
% The data of Electronics, Home, and Sports categories from both datasets serve as IND sources and the Tools category is used for the OOD test. 
% %
% Following the data processing protocol of UnisRec~\cite{hou2022towards}, we remove the products without metadata in the 2018 version dataset and conduct the 5-core filter to ensure all products and users appear at least 5 times. After filtering, we sort every user's interactions chronologically and truncate the history with a maximum of 50 products, retaining the least recent history.
%     %
% For the text information of products, we use the metadata from Amazon Review 2014 to fill in the missing information of the same products in the 2018 version metadata. 
% %
% We also combine the product title, category, and brand to represent a product. The average length of the combined texts is about 21 words. Thus, we retain the first, maximum of 25 words of each combined text for computational efficiency. 
% %following the protocol of UnisRec.
%     %
% As in conventional sequential recommendation tasks, we split the last product of the user interactions into the test set as the ground truth next product of the user's interest, the second last product into the validation set, and the remaining products into the training set.
%     %
% When processing the user interactions with instruction templates, for each sample, we randomly select 19 candidate products and mix them up with one ground-truth product. These 20 products serve as the options for the sample.

% \label{sec:appendix:data}

\section{Instruction Templates}
\label{sec:appendix:instr}
\subsection[]{Answerability Prediction (\AP)}
\paragraph{Captioning Instruction}
Please generate an informative caption for the product in the image. The caption should be helpful to identify if the product-related question: \{\{question\}\}, is answerable.
\paragraph{Caption Quality Evaluation Instruction}
The task needs to identify if the question is answerable based on the related document: \{\{review\}\}. Here is the additional information about the product that was extracted from the product image: \{\{caption\}\}. You need to determine if the information extracted from the image will help to identify the question's answerability. Only output yes or no.
\paragraph{Task Instruction}
Analyze the question and its supporting document, as well as the potential extra information about the products extracted from the product images, predict if the question is answerable based on the provided information. Output only yes or no.

\subsection[]{Category Classification (\CC)}
\paragraph{Captioning Instruction}
Please generate an informative caption for the product in the image. Here is the product title: \{\{title\}\}. The caption should be helpful in identifying the product's fine-grained category.
\paragraph{Caption Quality Evaluation Instruction}
The task needs to identify the product's fine-grained category from the options: \{\{options\}\}. Here is the additional information about the product that was extracted from the product image: \{\{caption\}\}. You need to determine if the information extracted from the image will help to identify the category. Only output yes or no.
\paragraph{Task Instruction}
Analyze the product title, as well as the potential extra information about the products extracted from the product images, identify the product category from the given options. Only answer from the options.

\subsection[]{Product Relation Prediction (\PRP)}
\paragraph{Captioning Instruction}
Please generate an informative caption for the product in the image. The title of the product in the image is \{\{title of the product\}\}. The caption should be helpful in predicting the relation between this product and \{\{title of another product\}\}.
\paragraph{Caption Quality Evaluation Instruction}
The model needs to identify if the two products are similar or will be purchased together or be viewed together given the title of product 1: \{\{title of the product\}\}, and product 2: \{\{title of another product\}\}. Here is the additional information about product 1 extracted from its image: \{\{caption of product 1\}\}, you need to determine if the information extracted from the image will be helpful in identifying the relation between the two products. Only output yes or no.
\paragraph{Task Instruction}
Given the title of two products, as well as the potential extra information about the products extracted from the product images, predict the relation of the two products. Only answer from the options.

\subsection[]{Product Substitute Identification (\PSI)}
\paragraph{Captioning Instruction}
Please generate an informative caption for the product in the image. The caption should be helpful to predict if the product: \{\{title\}\} can serve as a functional substitute for the user's query: \{\{query\}\}.
\paragraph{Caption Quality Evaluation Instruction}
The model needs to identify if the product is somewhat relevant to the query but fails to fulfill some aspects of the query but the product can be used as a functional substitute. Given a user's query: \{\{query\}\} and a product title: \{\{title\}\}, as well as additional information about the product extracted from the product image: \{\{caption\}\}, you need to determine if the information extracted from the image will be helpful in identifying the relevance between the product and the query. Only output yes or no.
\paragraph{Task Instruction}
Given a user's query and a product title, as well as the potential extra information about the product extracted from the product image, identify if the product is somewhat relevant to the query but fails to fulfill some aspects of the query but the product can be used as a functional substitute. Only output yes or no.

\subsection[]{Multi-class Product Classification (\MPC)}
\paragraph{Captioning Instruction}
Please generate an informative caption for the product in the image. The caption should be helpful to predict the relevance between the user's query: \{\{query\}\}, and product: \{\{title\}\}.
\paragraph{Caption Quality Evaluation Instruction}
The model needs to predict the relevance between the query and product by analyzing the user's query: \{\{query\}\}, and product title: \{\{title\}\}. Here is the additional information about the product extracted from the product image: \{\{caption\}\}, you need to determine if the information extracted from the image will be helpful in predicting the result. Only output yes or no.
\paragraph{Task Instruction}
Predict the relevance between the query and product by analyzing the user's query, and product title, as well as the potential extra information about the product extracted from the product image. Output the option that best describes the relevance.

\subsection[]{Sentiment Analysis (\SA)}
\paragraph{Captioning Instruction}
Please generate an informative caption for the product in the image. The caption should be helpful to identify the user's sentiment from the review: \{\{review\}\}.
\paragraph{Caption Quality Evaluation Instruction}
The task needs to identify the user's sentiment based on their review: \{\{review\}\}. Here is the additional information about the product extracted from the user review's image: \{\{caption\}\}. You need to determine if the information extracted from the image will help to identify the user's sentiment. Only output yes or no.
\paragraph{Task Instruction}
Given the user's review, as well as the potential extra information about the products extracted from the user review's image, identify the user's sentiment. Only answer from the options.

\subsection[]{Sequential Recommendation (\SR)}
\paragraph{Captioning Instruction}
Please generate an informative caption for the product in the image. Here is the product title: \{\{title\}\}. The caption should be helpful in predicting the next product the user is most likely to purchase by analyzing the user's intent based on the user's purchase history.
\paragraph{Caption Quality Evaluation Instruction}
The task needs to recommend the next product that the user may be interested in based on the user's purchase history. Here is the title of a product from purchase history: \{\{title, category, brand\}\}, and the information extracted from the product image: \{\{caption\}\}. You need to determine if the information extracted from the image will be helpful for recommendation. Only output yes or no.
\paragraph{Task Instruction}
Estimate the user's intent based on the user's purchase history, and predict the next product that the user is most likely to purchase from the given options.

\section[]{Analysis on \ECCC and \CQE}
\label{sec:appendix:analysis}
%=========================================================
In this section, we explore the impact of captioning models in \ECCC and caption quality evaluation models in \CQE on the performance of \pipeline, exemplified by \pipelineM.

%=========================================================
\subsection{Analysis on Captioning Models}
\label{sec:cap_sel}
%=========================================================

\begin{table*}[!t]
  % \xia{bold, formatting; confusing: captioning model vs no captions}}
  \centering
  % \vspace{-3pt}
  \footnotesize
  \begin{threeparttable}
      \begin{tabular}{
        @{\hspace{2pt}}c@{\hspace{2pt}}
        @{\hspace{2pt}}c@{\hspace{2pt}}
	  @{\hspace{2pt}}c@{\hspace{2pt}}
	  @{\hspace{2pt}}c@{\hspace{2pt}}%macrof1
	  @{\hspace{2pt}}c@{\hspace{2pt}}%f1
	  @{\hspace{2pt}}c@{\hspace{2pt}}%macrof1
	  @{\hspace{2pt}}c@{\hspace{2pt}}%hr1
        @{\hspace{2pt}}c@{\hspace{2pt}}%acc
        @{\hspace{2pt}}c@{\hspace{2pt}}%macrof1
        @{\hspace{2pt}}c@{\hspace{2pt}}%ndcg
        % @{\hspace{3pt}}c@{\hspace{2pt}}%impro
      }
      \toprule
      % \multicolumn{3}{c}{\multirow{2}{*}{Model}} 
      \multirow{2.5}{*}{\textbf{Model}} & \multirow{2.5}{*}{\textbf{Setting}} & \multirow{2.5}{*}{\textbf{Captioning Model}}
      % & \multirow{2}{*}{Inspector}
      & \AP & \CC & \PRP & \PSI & \MPC & \SA & \SR \\ % & \multirow{2}{*}{impro(\%)}\\
      \cmidrule{4-10}
      &&& F1 & R@1 & M-F1 & F1 & Acc & M-F1 & R@1 \\

      \midrule 
      
      \texttt{ft}-LLaVA-NExT-Interleave & w image & - & 0.791 & 0.964 & 0.568 & 0.340 & 0.721 & 0.561 & 0.053 \\
      \texttt{ft}-LLaVA-NExT-Interleave$^*$ & w caption & Llama-3.2-Vision-Instruct
       & 0.633 & 0.961 & 0.552 & 0.404 & 0.722 & 0.579 & 0.000 \\
      
      \midrule
      
      % \texttt{ft}-Mistral-7B-v0.3
      \uniMMM-$\mathtt{M}$ & w/o caption & -
      & 0.876 & 0.971 & 0.533 & 0.312 & 0.725 & 0.617 & 0.218 \\
      
      \midrule

      \multirow{5}{*}{\pipelineM} 

      & \multirow{2}{*}{w/o context} & BLIP2-OPT-2.7B
      & 0.878 & 0.976 & 0.545 & 0.352 & \textbf{0.734} & 0.614 & 0.209 \\ 
      && Llama-3.2-Vision-Instruct & 0.880 & 0.978 & 0.520 & 0.392 & 0.727 & 0.633 & 0.214 \\ 
      \cmidrule{2-10}
      
      & \multirow{3}{*}{w/ context \& caption}
      &  LLaVA-1.5-7B
      & 0.886 & \textbf{0.987} & 0.532 & 0.450 & 0.725 & 0.637 & 0.213 \\ 

      && LLaVA-NExT-mistral-7B
      & 0.886 & 0.979 & 0.558 & \textbf{0.476} & 0.725 & 0.647 & 0.210  \\

      && Llama-3.2-Vision-Instruct
      & \textbf{0.891} & 0.979 & \textbf{0.566} & 0.398 & 0.731 & \textbf{0.656} & \textbf{0.223} \\
      
      % \midrule

      % \multirow{9}{*}{\methodS} 
      
      % & \multirow{2}{*}{Blip2} % blip2-opt-2.7b
      % & \allcap & 0.840 & 0.941 & 0.494 & 0.000 & 0.683 & 0.565 & 0.160 \\
      % && \majvote & 0.839 & 0.945 & 0.491 & 0.009 & 0.686 & 0.571 & 0.169 \\ 
      % \cmidrule{2-10}
      
      % & \multirow{2}{*}{LLaVA} % llava-1.5-7b-hf
      % & \allcap & 0.834 & 0.943 & 0.475 & 0.017 & 0.698 & 0.593 & 0.155 \\
      % && \majvote & 0.843 & 0.941 & 0.487 & 0.000 & 0.692 & 0.546 & 0.173 \\ 
      % \cmidrule{2-10}
      
      % & \multirow{2}{*}{LLaVA-NExT} % llava-v1.6-mistral-7b-hf
      % & \allcap & 0.790 & 0.947 & 0.495 & 0.034 & 0.701 & 0.523 & 0.081 \\
      % && \majvote & 0.764 & \textbf{0.953} & 0.485 & 0.026 & 0.695 & 0.547 & 0.170  \\ 
      % \cmidrule{2-10}
      
      % & \multirow{2}{*}{Llama-3.2-Vision} 
      % & \allcap & 0.838 & 0.943 & 0.489 & 0.035 & 0.702 & 0.583 & 0.139 \\
      % && \majvote & \textbf{0.848} & 0.941 & \textbf{0.496} & \textbf{0.059} & \textbf{0.713} & \textbf{0.598} & \textbf{0.171} \\ 

      \bottomrule
      \end{tabular}
      % \begin{tablenotes}[normal, flushleft]
      % \begin{footnotesize}
      % \item
      
      % \par
      % \end{footnotesize}
      % %\end{scriptsize}
      % \end{tablenotes}
  % \vspace{-10pt}
  \end{threeparttable}
  \caption{Comparison using Different Captioning Models. The best performance on each task is in \textbf{bold}. When employing different caption models, we only involve captions that are predicted to be useful by \CQE. $^*$ indicates the version of LLaVA-NExT-Interleave fine-tuned and evaluated on captioning data generated by \ECCC and \CQE.}
  \label{tbl:captioner}
\end{table*}

%\label{tbl:captioner}

When analysis the impact of captioning models,
we include BLIP2-OPT-2.7B~\cite{li2023blip2} as a context-free captioning model and evaluate it as a baseline.
Table~\ref{tbl:captioner} also compares the \pipelineM using various individual 
captioning models, including
% Blip2~\cite{li2023blip2}, 
LLaVA-1.5-7B~\cite{liu2023visual,liu2024improved}, 
LLaVA-NExT-mistral-7B~\cite{liu2024llavanext}, 
and Llama-3.2-Vision-Instruct~\cite{dubey2024llama3}. 
%We also use Blip2~\cite{li2023blip2} for context-free caption generation.
%\xy{I think maybe it is enough in abolition study}
% \xia{ordering!}
%\xia{need these full names in table 4 and also in this section.}
Table~\ref{tbl:captioner} presents the results. 

%
%Overall, \textbf{\textit{the use of different captioning models in the context-conditioned setting reach comparable performances that are all better than when no captions or context-free captions are used}}
%
%\xy{Overall, \textbf{\textit{the use of different captioning models 
%in the context-conditioned setting reach superior performances that are all better than when no captions or context-free captions are used. In addition, employing different captioning models yields comparable results while using Llama-3.2-Vision-Instruct yields consistently better results}}. }

% \xia{need to rewrite this section in this order: 
% 1. using context-conditioned captions is overall better than 
% no captions; 2. usng context-conditioned captions is overall better than context-free captions; 
% 3. among all context-conditioned captioning methods, comparable results but the last one 
% is overall slightly and consistently better.}

% \xia{rewrite based on the analysis below}

\textbf{(1)}
\textbf{\textit{Overall, using visual information through captioning
is almost always better than not using visual information. }}
Specifically, 
using BLIP2-OPT-2.7B to generate context-free captions from images brings a 1.8\% average improvement compared with 
% \textit{ft}-Mistral-7B-v0.3
\uniMMM-$\mathtt{M}$
, which does not use visual information at all; 
%using LLaVA as the captioning model brings 6.8\% improvement 
%compared with \textit{ft}-Mistral-7B-v0.3, which does not use visual information at all;
% 
using LLaVA-NExT-mistral-7B in \pipeline for context-conditioned captioning results in 8.6\% improvement over 
% \textit{ft}-Mistral-7B-v0.3
\uniMMM-$\mathtt{M}$
.  
%and Llama-3.2-Vision-Instruct delivers 6.5\% improvement. 
%
This shows the utility of visual information in e-commerce tasks and demonstrates that 
captioning is an effective way of utilizing images in e-commerce models. 

%These improvements demonstrates the robustness of \pipeline, as the choice of captioning models 
%% has a similar positive effect across the board. No matter what captioning model is used, \pipeline consistently enhances the performance.
%consistently yields positive results. 
%Regardless of which captioning model is used, \pipeline reliably enhances performance, 
%reinforcing its effectiveness across the board. 

% \xy{Meanwhile, using the context-free captioning model (Blip2) as the captioning model only brings 1.8\% improvement over \textit{ft}-Mistral-7B-v0.3.
\textbf{(2)}
\textbf{\textit{Context-condition captioning beats context-free captioning for e-commerce}}. 
\pipelineM, which employs Llama-3.2-Vision-Instruct as the captioning model by default, outperforms that using the context-free captioning model (BLIP2-OPT-2.7B) by 4.5\%. This further highlights the advantage of using context-conditioned captioning to enhance task performance compared to more generic, context-free approaches.
Comparing all context-conditioned captioning models, we observe comparable results, but 
\textbf{\textit{Llama-3.2-Vision-Instruct as the captioning model is slightly and consistently better overall}}. 

\textbf{(3)}
\textbf{\textit{\pipeline is possessed with better capability leveraging captions than MFM}}. 
\texttt{ft}-LLaVA-NExT-Interleave using captions for the text input improves \AP and \PSI slightly compared to its image-using counterpart. However, this approach falls behind \pipelineM across most tasks. This indicates that using captions as a substitute for the original multimodal input in MFMs is suboptimal. MFMs are designed to process multimodal inputs directly, leveraging both visual and textual modalities simultaneously, and are not fully optimized for text-only inputs. The results underscore that simply incorporating captions into MFMs is insufficient to fully leverage the multimodal information cohesively and effectively.

%=========================================================
\subsection{Analysis on Evaluation Strategies}
\label{sec:eval_sel}
%=========================================================

% \xia{this section needs rewrting: use the terms we agreed upon. it is not clear what these methods are used for.}

\begin{table*}
  % \xia{formatting; one column; full name}}
  \centering
  % \vspace{-3pt}
  \footnotesize
  \begin{threeparttable}
      \begin{tabular}{
        @{\hspace{4pt}}c@{\hspace{4pt}}
        @{\hspace{4pt}}c@{\hspace{6pt}}
	  @{\hspace{4pt}}c@{\hspace{3pt}}%macrof1
	  @{\hspace{3pt}}c@{\hspace{3pt}}%f1
	  @{\hspace{3pt}}c@{\hspace{3pt}}%macrof1
	  @{\hspace{3pt}}c@{\hspace{3pt}}%hr1
        @{\hspace{3pt}}c@{\hspace{3pt}}%acc
        @{\hspace{3pt}}c@{\hspace{3pt}}%macrof1
        @{\hspace{3pt}}c@{\hspace{3pt}}%ndcg
        % @{\hspace{3pt}}c@{\hspace{2pt}}%impro
      }
      \toprule
      % \multicolumn{3}{c}{\multirow{2}{*}{Model}} 
      \multirow{2}{*}{\textbf{Strategy}} & \multirow{2}{*}{\textbf{Evaluation Model}}
      & \AP & \CC & \PRP & \PSI & \MPC & \SA & \SR \\ % & \multirow{2}{*}{impro(\%)}\\
      \cmidrule(lr){3-9}
      && F1 & R@1 & M-F1 & F1 & Acc & M-F1 & R@1 \\    
      \midrule

      \allcap & - & 0.885 & 0.976 & 0.535 & 0.352 & 0.722 & 0.642 & 0.207 \\
      \midrule
      
      \multirow{5}{*}{Single}
      
      & Llama-3.2-3B-Instruct & 0.884 & 0.971 & 0.512 & 0.395 & 0.731 & 0.603 & 0.216 \\
      & Phi-3.5-mini-Instruct & 0.885 & 0.976 & 0.515 & 0.294 & 0.733 & 0.638 & 0.210 \\
      & Mistral-7B-Instruct-v0.3 & 0.879 & 0.976 & 0.540 & 0.389 & 0.737 & 0.651 & 0.212 \\
      & Llama-3.1-8B-Instruct & 0.885 & 0.974 & 0.549 & 0.404 & 0.722 & 0.622 & 0.220 \\
      & Llama-3.2-Vision-Instruct & 0.885 & 0.969 & 0.538 & 0.397 & 0.737 & 0.622 & \textbf{0.223} \\

      \midrule
      
      \multirow{3}{*}{\majvote} 
      & 3 models & 0.881 & 0.969 & 0.543 & 0.396 & 0.719 & 0.631 & 0.218 \\
      & 5 models & \textbf{0.891} & 0.979 & \textbf{0.566} & 0.398 & 0.731 & 0.656 & \textbf{0.223} \\
      & 7 models & 0.882 & \textbf{0.984} & 0.546 & \textbf{0.416} & \textbf{0.740} & \textbf{0.659} & 0.219 \\
      % & {\color{brown}selected ratio (\%)} & {\color{brown}75.6} & {\color{brown}23.7} & {\color{brown}50.5} & {\color{brown}20.9} & {\color{brown}38.0} & {\color{brown}68.8} & {\color{brown}28.6} \\

      \bottomrule
      \end{tabular}
      % \begin{tablenotes}[normal, flushleft]
      % \begin{footnotesize}
      % \item
      % \par
      % \end{footnotesize}
      % %\end{scriptsize}
      % \end{tablenotes}
 % \vspace{-10pt}
  \end{threeparttable}
  \caption{Comparison of Caption Quality Evaluation Methods in IND Evaluation. 
      The best performance on each task is in \textbf{bold}. The results are evaluated from \pipelineM.}
  \label{tbl:inspector}
\end{table*}

% \label{tbl:inspector}

In Table~\ref{tbl:inspector}, we compare \pipelineM using different caption quality evaluation strategies, including using a single evaluation model, and 
majority voting (\majvote) from 3, 5, and 7 models. 
For majority voting with 3 CQE models, we use Llama-3.1-8B-instruct, Llama-3.2-vision-instruct, and Mistral-7B-instruct-v0.3 as evaluation models. 
For five-model voting, we added Phi-3.5-mini-instruct and Llama-3.2-3B-instruct as evaluation models. 
For seven-model voting, we further include Llama-3-8B-instruct and qwen2.5-7B-instruct as evaluation models.
We also compare the strategy when the caption is used always (i.e., \allcap), 
all with Llama-3.2-Vision-Instruct serving as the captioning model (\ECCC).

\textbf{(1)}
\textbf{\textit{Compared with \allcap, using caption quality evaluation models brings 
performance improvement in general.}}
%
%Specifically, the five individual strategies outperform \allcap at 1.0\%, -2.6\%, 2.4\%, 2.9\%, 2.8\%
%\xia{\hl{I stopped here. need to discuss}!}
% \bo{\st{The first five rows show the performance of using each specific model individually as the evaluation model, bringing varying degrees of improvement (1.0\%, -2.6\%, 2.4\%, 2.9\%, 2.8\% respectively compared with \allcap).}}
%
%\bo{In Table~\ref{tbl:inspector}, the middle five rows show the performance of each single model when used as the evaluator, 
%
%As shown in Table~\ref{tbl:inspector}, compared to \allcap, using a single evaluation model leads to a considerable average improvement of up to 2.9\% over the 7 tasks.}
%
As shown in Table~\ref{tbl:inspector}, compared to \allcap, using all evaluation models together with \majvote leads to a considerable average improvement of 4.4\%.

\textit{
\textbf{(2)}
\textbf{Compared to using a single evaluation model, \majvote-based evaluation leads to further improvement.}
}
Notably, employing \majvote-based evaluation, 
which combines the results of all evaluation models, 
%yields higher improvements (4.4\% compared with \pipelineM-\allcap) 
%than using a single evaluation model, 
yields higher performance 
than using a single evaluation model (1.7\% improvement over \pipelineM with Llama-3.2-Vision-Instruct as the evaluation model)
highlighting the effectiveness of our \majvote evaluation strategy.
%
%\bo{yields additional improvement compared to that using a single evaluation model.
%
%For example, XXX.
%
%This result highlights the effectiveness of our \majvote evaluation strategy.
%
% \bo{@Xinyi you can compare \majvote with \allcap directly.}
% \xy{we already compared that in [Comparison Between Using \majvote and \allcap] in \ref{sec:exp_results:mv_allcap}}
% \xia{still need to briefly compare \majvote and \allcap}
% \paragraph{\st{Comparison Between Using {\majvote} and {\allcap}}}
% \label{sec:exp_results:mv_allcap}
% / Comparison Between \pipelineM and \pipelineM-\allcap

% \textbf{(2)}
% \textbf{\textit{Deliberately selecting captions (i.e., \majvote) results in a 6.5\% improvement 
% (\pipelineM vs \textit{ft}-Mistral-7B-v0.3) 
% % (\pipeline with \methodM-\majvote vs \textit{ft}-Mistral-7B-v0.3) 
% across all 7 tasks. }}
% %
% When comparing \pipeline's use of the \allcap and \majvote evaluation strategies, 
% we observe that indiscriminately incorporating all captions 
% leads to only a 1.9\% improvement over using text-only information 
% (\pipelineM-\allcap vs \textit{ft}-Mistral-7B-v0.3) vs \majvote's 6.5\%. 
% \bo{@Xinyi I do not follow here. Do not think we need it.}
%
\textit{
\textbf{(3)}
\textbf{Compared to using a various number of evaluation models by \majvote, five evaluation models yield the comparable high performance with less cost.}}
Specifically, incorporating five evaluation models yields a 2.1\% average improvement compared to three models. However, increasing to seven evaluation models provides only a marginal 0.1\% improvement over five models. To balance computational cost and performance, we opted to use five models in the \CQE module. The results offer deeper insights into the framework's design choices and substantiate our approach.

%=========================================================
\subsection{Analysis on Context-conditioned Captions}
\label{sec:appendix:ccc}
%=========================================================
While some overlap is natural since both captions and titles describe product attributes, our \ECCC module generates context-conditioned captions that go beyond static title information. Unlike titles, which are often short, seller-centric, and lack contextual adaptation, \ECCC enriches captions with task-relevant visual evidence. For example, in Figure~\ref{fig:case_mpc}, given a user query highlighting “wings”, \ECCC produces “A Labrador Retriever dressed as a yellow angel with moving wings, designed as a tree topper”, which captures fine-grained, query-relevant visual details absent in the product title.

To quantify overlap, we conducted a systematic analysis of generated captions and product titles. We calculate the Jaccard similarity, which computes the percentage of word overlap between two sentences, and the semantic similarity, which calculates the cosine similarity of two sentences’ embedding. The results are demonstrated in Table~\ref{tbl:similarity}.

\begin{table*}
  \centering
  \footnotesize
  \begin{threeparttable}
      \begin{tabular}{
        @{\hspace{4pt}}c@{\hspace{4pt}}
	  @{\hspace{4pt}}r@{\hspace{3pt}}%macrof1
	  @{\hspace{3pt}}r@{\hspace{3pt}}%f1
	  @{\hspace{3pt}}r@{\hspace{3pt}}%macrof1
	  @{\hspace{3pt}}r@{\hspace{3pt}}%hr1
        @{\hspace{3pt}}r@{\hspace{3pt}}%acc
        @{\hspace{3pt}}r@{\hspace{3pt}}%macrof1
        @{\hspace{3pt}}r@{\hspace{5pt}}%ndcg
      }
      \toprule
      Similarity & \AP & \CC & \PRP & \PSI & \MPC & \SA & \SR \\  
      \midrule
      Jaccard Similarity (\%) & 6.51 & 17.00 & 15.51 & 15.84 & 16.67 & 14.36 & 15.75 \\
      Semantic Similarity (\%) & 46.36 & 75.89 & 77.35 & 70.91 & 73.17 & 57.65 & 75.38 \\

      \bottomrule
      \end{tabular}
  \end{threeparttable}
  \caption{Caption-title Similarity}
  \label{tbl:similarity}
\end{table*}

The very low Jaccard similarity scores confirm limited word-level overlap, while the higher semantic similarity reflects that both describe the same product but from complementary perspectives. Crucially, captions highlight visual grounding (e.g., colors, arrangements, subtle details) that titles do not encode. Empirically, our ablations (Table~\ref{tbl:ablation}, \uniMMM vs. \pipeline) demonstrate that unimodal fine-tuning on titles alone cannot match \pipeline’s performance, validating that captions provide distinctive and non-redundant contributions.

%=========================================================
\subsection{Real-world Considerations}
\label{sec:appendix:real-world}
%=========================================================
When considering the real-world situation, scalability, computational costs, or integration in environments are important.
\pipeline is inherently deployable, as it avoids joint end-to-end multimodal training. Take the \MPC task as an example, we calculate the runtime of \CQE with 5 LLMs and result in ~0.4s per instance since each model only needs to answer yes-no questions. Besides, \pipeline's modularity allows seamless substitution or refinement of components in e-commerce environments.
%\label{sec:appendix:abolition}

%\section{Experimental Details}
%\label{sec:appendix:full_results}
%In this section, we demonstrate the detailed experimental setup and results as supplementary.

\section{Detailed Experimental Setup}
\label{sec:appendix:exp_setup}

\paragraph{Fine-tuned CLIP-based Models}
FashionCLIP~\cite{chia2022fashionclip} is 
a SoTA CLIP-based~\cite{radford2021clip} model 
adapted to the \mbox{e-commerce} fashion domain 
and is skilled at various multimodal tasks. 
We fine-tune the Huggingface checkpoint of 
FashionCLIP on each task using the \dataset training set
and denoted the fine-tuned model as \texttt{ft}-FashionCLIP.
% and evaluate the fine-tuned model on the \dataset test sets.

% \paragraph{Fine-tuned LLMs}
% % \xy{(1. use 4 models; 2. fine-tune 3 general LLMs; 3. eCeLLM.)}
% %
% We use 3 LLMs as the baselines. 
% For Llama-2-13b-chat~\cite{touvron2023llama2}, 
% Mistral-7B-Instruct-v0.3~\cite{jiang2023mistral}, 
% and Llama-3.2-3B-Instruct~\cite{dubey2024llama3}, 
% we fine-tune their checkpoints  
% released in Huggingface on \dataset training data
% using all tasks and only text as input. 
% The \mbox{fine-tuned} models are denoted as 
% \texttt{ft}-Llama-2-13B, \texttt{ft}-mistral-7B-v0.3, and \texttt{ft}-Llama-3.2-3B.
% % \hanwen{this format might look better: Llama-2-13B-\texttt{ft}}, respectively.
% % \xia{update the tables accordingly}
% % We conduct the zero-shot evaluation on the fine-tuned model. 
% We perform the zero-shot evaluation on 
% the fine-tuned models 
% since these models have already gained e-commerce knowledge.

\paragraph{Fine-tuned MFMs}
We fine-tune LLaVA-NExT-interleave-qwen-7b~\mbox{\cite{li2024llava-next-interleave}} as the MFM baseline, 
% \hl{On the one hand,} \xia{weird wording?} \hl{e-commerce applications may involve multiple products. } \xia{this is not accurate and is not the same thing as in your example} 
% For example, the \PRP task evaluates the relation between two products. \hl{On the other hand} \xia{I don't understand what the following (multiple images) is the ``other hand" -- what are you contrasting?}, a product typically has one or more images in most e-commerce \hl{scenarios} \xia{not accurate}.
% 
% \hl{Our {\dataset} needs models to deal with one or multiple products simultaneously.
% For example, the {\PRP} task evaluates the relation between two products.
% Therefore, a multi-image multimodal foundation model such as 
% LLaVA-NExT-interleave-qwen-7b is applicable to evaluate e-commerce tasks.}
%
which 
is a SoTA multi-image MFM 
able to process input
textual and image information of one or multiple instances, making it a suitable baseline 
for e-commerce tasks, particularly those evaluating 
multiple products simultaneously (e.g., \PRP).
We fine-tune the checkpoint of LLaVA-NExT-interleave-qwen-7b 
released in Huggingface on the training data of \dataset.
The fine-tuned model is denoted as \texttt{ft}-LLaVA-NExT-interleave.
We also conduct the zero-shot evaluation for this baseline. 
%as the model already 
%possesses a comprehensive understanding of e-commerce concepts.
% \xia{the last sentence is out of nowhere and not connected to the previous one. }

\paragraph{E-commerce LLMs}
We utilize eCeLLM-L and eCeLLM-M~\cite{peng2024ecellm}, 
a series of SoTA e-commerce LLMs,
fine-tuned on various e-commerce tasks, as a baseline. 
For \mbox{eCeLLM-L} and \mbox{eCeLLM-M}, we perform a zero-shot evaluation 
using the checkpoints available on Huggingface 
since they already encompass a broad understanding of e-commerce concepts.
% \hanwen{probably rationalize why we use the middle size model? Reviewers may think: you are afraid that you cannot beat the largest model?}

\paragraph{SoTA Task-Specific Models}
To evaluate the \SR and \CC tasks,  
we fine-tune \mbox{Recformer}~\cite{li2023recformer}, 
a popular language-based recommendation model, 
and Sentence-BERT~\cite{reimers2019sbert}, 
which is adept at semantic similarity search tasks like retrieval, respectively.
All other tasks are evaluated on the 
fine-tuned DeBERTa~\cite{he2021deberta}, 
which is a widely used BERT-based model known for 
its strong performance in various language understanding tasks. 
%and a solid choice for achieving competitive results. 

\paragraph{Hyperparameters and Reproducibility}
The learning rate and batch size are set as 1e-4 and 128 during fine-tuning of all the models. A cosine learning rate scheduler with a 5\% warm-up period for 3 epochs is applied. We set $\alpha$ and the rank in LoRA as 16, and add LoRA adaptors to all the projection layers and the language modeling head. 
We perform zero-shot evaluations (i.e., without in-context examples) on all the tasks.
\section{Detailed Experimental Results}
\label{sec:appendix:full_results}
%In this section, we demonstrate the detailed experimental setup and results as supplementary.
\subsection{More IND Results}
\label{sec:appendix:ind_result}
In this section, we bring more discussion on in-domain (IND) evaluation as a supplementary of Section~\ref{sec:ind} from Table~\ref{tbl:performance_ind}.

\textbf{(1)} 
\textbf{\textit{\pipelineM achieves a significant 45.8\%
improvement over the \textit{ft}-FashionCLIP fine-tuned on the training data of \dataset}}.
A key difference between \pipeline and FashionCLIP is that 
\pipeline uses the textual representation of images generated via context-conditioned captioning (\ECCC), 
adjusting the focus on image details with respect to the specific context. 
In contrast, FashionCLIP generates image representations without considering the specific context.
Additionally, \pipeline could leverage the extensive world knowledge of LLMs to enrich the captions, while FashionCLIP considers the images solely using the vision encoder.

\textbf{(2)}
\textbf{\textit{\pipelineM outperforms SoTA \mbox{task-specific} models with a significant 22.1\% 
improvement across all 7 tasks.}}
Compared with SoTA task-specific models, 
which solely rely on textual information from each individual task, 
\pipeline could leverage both vision and language information 
of each task, and the information shared across diverse e-commerce tasks, 
as well as LLM's inherent knowledge and learning power, 
to significantly boost performance on each individual task.

\textbf{(3)}
\textbf{\textit{Mid-size \pipelineM performs best among \pipeline model sizes.}}
% This conforms to the findings in the literature~\cite{peng2024ecellm} 
Benefitting from the large-scale instruction-tuning dataset and powerful base model (Mistral-7B-Instruct-v0.3)
% When the instruction-tuning
% dataset is of large scale but not as large as the training data of foundation models, 
mid-size fine-tuned models achieve most, balancing learning from instruction tuning while 
retaining knowledge from base models.

\textbf{(4)}
Considering the percentage of captions selected by \majvote, \textbf{\textit{sparse caption usage still leads to high gains, implying a strong signal when captions are selected}}. For example, \SR only uses captions 30\% of the time but leads an 18.6\% gain in IND evaluation.

%=========================================================
\subsection[]{Comparison with Proprietary Models}
\label{sec:appendix:close}
%=========================================================
We have conducted new experiments with Claude-3.5 and GPT-4o (both text-only and multimodal) to \dataset against our proposed \pipeline models (-S, -M, -L). Evaluation results on IND and OOD test sets are summarized in Table~\ref{tbl:performance_close}.

\begin{table*}[htbp]
  \centering
  \begin{footnotesize}
  \begin{threeparttable}
      \begin{tabular}{
        @{\hspace{8pt}}l@{\hspace{4pt}}
	  @{\hspace{4pt}}c@{\hspace{4pt}}%macrof1
	  @{\hspace{4pt}}c@{\hspace{4pt}}%f1
	  @{\hspace{4pt}}c@{\hspace{4pt}}%macrof1
	  @{\hspace{4pt}}c@{\hspace{4pt}}%hr1
	  @{\hspace{4pt}}c@{\hspace{4pt}}%f1
	  @{\hspace{4pt}}c@{\hspace{4pt}}%macrof1
	  @{\hspace{4pt}}c@{\hspace{0pt}}%f1
        @{\hspace{4pt}}c@{\hspace{4pt}}%None
	  @{\hspace{4pt}}c@{\hspace{4pt}}%macrof1
	  @{\hspace{4pt}}c@{\hspace{4pt}}%hr1
	  @{\hspace{4pt}}c@{\hspace{4pt}}%f1
	  @{\hspace{4pt}}c@{\hspace{4pt}}%macrof1
	  @{\hspace{4pt}}c@{\hspace{8pt}}%macrof1
      }
      \toprule
      \multicolumn{1}{c}{\multirow{2.5}{*}{\textbf{Model}}} 
      & \multicolumn{7}{c}{\textbf{IND}} & & \multicolumn{5}{c}{\textbf{
      OOD}} \\
      \cmidrule{2-8} \cmidrule{10-14}
      & \AP & \CC & \PRP & \PSI & \MPC & \SA & \SR & & \AP & \CC & \PRP & \SA & \SR \\
      \midrule
          
      Claude-3.5 (text-only)    & 0.656 & 0.949 & 0.329 & 0.360 & 0.683 & 0.433 & 0.107 & & 0.659 & 0.974 & 0.259 & 0.427 & 0.196 \\
      Claude-3.5   & 0.755 & 0.952 & 0.360 & 0.366 & 0.657 & 0.481 & 0.069 & & 0.752 & 0.976 & 0.255 & 0.470 & 0.139 \\
      GPT-4o (text-only)        & 0.475 & 0.971 & 0.401 & 0.315 & 0.581 & 0.525 & 0.151 & & 0.487 & \textbf{0.989} & 0.319 & 0.539 & 0.213 \\
      GPT-4o       & 0.510 & \textbf{0.982} & 0.441 & 0.252 & 0.664 & 0.538 & 0.123 & & 0.551 & 0.987 & 0.331 & 0.565 & 0.217 \\

      \midrule
      \pipelineL & 0.868 & 0.969 & 0.473 & 0.268 & 0.706 & 0.651 & 0.190
      && 0.840 & 0.968 & 0.531 & 0.607 & 0.297\\ 
      \pipelineM
      & \textbf{0.891} & 0.979 & \textbf{0.566} & \textbf{0.398} & \textbf{0.731} & \textbf{0.656} & \textbf{0.223} 
      && 0.855 & 0.977 & \textbf{0.585} & 0.625 & \textbf{0.330} \\
      
      \pipelineS & 0.871 & 0.963 & 0.504 & 0.336 & 0.707 & 0.601 & 0.196
      && \textbf{0.857} & 0.959 & 0.580 & \textbf{0.647} & 0.297 \\

      \bottomrule
      \end{tabular}
  \end{threeparttable}
  \end{footnotesize}
  %\vspace{-3pt}
  \caption{Performance Comparison with Proprietary Models. The best performance on each task is in \textbf{bold}}
  \label{tbl:performance_close}
  %\vspace{-8pt}
\end{table*}

% \label{performance_close}

As shown, \pipelineM consistently outperforms both Claude-3.5 and GPT-4o across nearly all tasks.
Under the IND setting, \pipelineM achieves the highest overall performance, with particularly large margins on \PRP (0.566) and \SR (0.223), surpassing GPT-4o (0.441 \PRP, 0.123 \SR) and Claude-3.5 (0.360 \PRP, 0.069 \SR). This trend remains consistent in the OOD setting, where \pipelineM obtains strong generalization. These improvements are particularly pronounced on complex reasoning tasks which require nuanced understanding of contextual and causal relationships.

Furthermore, other variants (\pipelineS and \pipelineL) also exhibit competitive or superior performance to both baselines in most metrics, demonstrating the robustness and scalability of the \pipeline architecture. 
Overall, these results highlight \pipeline’s competitiveness against advanced proprietary models, affirming its strong adaptability and reasoning ability across diverse visual-linguistic domains.

%=========================================================
\subsection[]{Error Analysis}
\label{sec:appendix:error}
%=========================================================
We conduct an error analysis with both taxonomy and quantification in using the captions as the visual representation in \pipelineM by sampling 100 failure cases. The observed errors are categorized into five error types:

\textbf{(1) Attribute missing} (18\%): image provides a specific attribute, but the caption fails to capture it.

\textbf{(2) Attribute hallucination} (7\%): caption introduces attributes not grounded in the image.

\textbf{(3) Context conflict} (31\%): useful product information is diluted or distracted by noisy visual details.

\textbf{(4) Helpful caption missing} (10\%): beneficial captions are incorrectly filtered out by CQE.

\textbf{(5) Hard cases} (34\%): captions are accurate, but the task itself is inherently difficult.

Across tasks, we find that context conflict and hard cases dominate. This taxonomy not only clarifies \pipeline’s failure modes but also points to actionable directions: refining caption prompts to reduce missing attributes, improving \CQE filtering to recover helpful captions, and exploring debiasing strategies to mitigate context conflicts.

\subsection{Detailed Results for All the Tasks}
\label{sec:appendix:exp_result}

\begin{table*}
  \centering
  %\vspace{-10pt}
  \footnotesize
  \begin{threeparttable}
      \begin{tabular}{
        @{\hspace{4pt}}l@{\hspace{4pt}}
	  @{\hspace{2pt}}l@{\hspace{4pt}}
	  @{\hspace{4pt}}c@{\hspace{4pt}}% Acc
	  @{\hspace{6pt}}c@{\hspace{4pt}}% M-Rec
	  @{\hspace{4pt}}c@{\hspace{5pt}}% M-Pre
	  @{\hspace{6pt}}c@{\hspace{2pt}}% M-F1
	  @{\hspace{1pt}}c@{\hspace{0pt}}% failed
        @{\hspace{4pt}}c@{\hspace{6pt}}%None
	  %@{\hspace{2pt}}c@{\hspace{2pt}}%None
	  %@{\hspace{2pt}}c@{\hspace{8pt}}%f1
	  @{\hspace{4pt}}c@{\hspace{4pt}}%macrof1
	  @{\hspace{6pt}}c@{\hspace{4pt}}%hr1
	  @{\hspace{4pt}}c@{\hspace{5pt}}%f1
	  @{\hspace{6pt}}c@{\hspace{2pt}}%macrof1
	  @{\hspace{1pt}}c@{\hspace{4pt}}%macrof1
      }
      \toprule
      \multicolumn{2}{c}{\multirow{3}{*}{\textbf{Model}}} 
      % \multirow{2}{*}{Model} 
      % & \multirow{2}{*}{Captioner} 
      % & \multirow{2}{*}{Inspector}
      & \multicolumn{5}{c}{\textbf{IND}} & & \multicolumn{5}{c}{\textbf{OOD}} \\
      \cmidrule{3-7} \cmidrule{9-13}
      %&& \AP & \CC & \PRP & \PSI & \MPC & \SA & \SR & \AP & \CC & \PRP & \SA & \SR \\
      %\cmidrule(lr){3-14}
      && Acc & M-Rec & M-Pre & M-F1 & \#Failed && Acc & M-Rec & M-Pre & M-F1 & \#Failed \\
      \midrule
      
      \multicolumn{2}{l}{\textit{ft}-LLaVA-NExT-Interleave} & 0.746 & 0.895 & 0.709 & 0.791 & 11 && 0.509 & 0.626 & 0.538 & 0.579 & 13\\
      \cmidrule{3-7} \cmidrule{9-13}

      \multicolumn{2}{l}{eCeLLM-L} & 0.821 & 0.851 & 0.894 & 0.872 & 0 && 0.814 & 0.813 & 0.912 & \textbf{0.860} & 0\\
      \multicolumn{2}{l}{eCeLLM-M} & 0.817 & 0.876 & 0.852 & 0.864 & 0 && 0.793 & 0.809 & 0.877 & 0.841 & 0\\
      
      \cmidrule{3-7} \cmidrule{9-13}
      
      \multicolumn{2}{l}{\textit{ft}-FashionCLIP} & 0.673 & 0.764 & 0.754 & 0.759 & 0 && 0.550 & 0.677 & 0.538 & 0.600 & 0\\
      \cmidrule{3-7} \cmidrule{9-13}
      
      \multicolumn{2}{l}{Task-specific Model} & 0.832 & \textbf{0.939} & 0.806 & 0.868 & 0 && \textbf{0.824} & \textbf{0.917} & 0.791 & 0.849 & 0\\

      \midrule

      \multirow{3}{*}{\pipelineL}
      & \uniMMM & 0.809 & 0.832 & 0.902 & 0.866 & 0 && 0.767 & 0.760 & 0.917 & 0.831 & 0\\
      & \ECCC-\uniMMM & 0.799 & 0.823 & 0.899 & 0.859 & 0 && 0.781 & 0.773 & 0.920 & 0.840 & 0\\
      & \ECCC-\CQE-\uniMMM & 0.812 & 0.833 & 0.906 & 0.868 & 0 && 0.782 & 0.776 & 0.915 & 0.840 & 0\\ 
      \midrule

      \multirow{3}{*}{\pipelineM} 
      & \uniMMM & 0.823 & 0.837 & 0.919 & 0.876 & 0 && 0.795 & 0.795 & 0.906 & 0.847 & 0\\
      & \ECCC-\uniMMM & 0.840 & 0.866 & 0.906 & 0.885 & 0 && 0.815 & 0.820 & 0.903 & 0.859 & 0\\
      & \ECCC-\CQE-\uniMMM & \textbf{0.846} & 0.863 & \textbf{0.921} & \textbf{0.891} & 0 && 0.813 & 0.831 & 0.880 & 0.855 & 0\\
      
      \midrule

      \multirow{3}{*}{\pipelineS}
      & \uniMMM & 0.808 & 0.825 & 0.912 & 0.866 & 0 && 0.772 & 0.756 & 0.939 & 0.838 & 0\\
      & \pipelineS-\allcap & 0.815 & 0.838 & 0.903 & 0.869 & 0 && 0.806 & 0.798 & 0.923 & 0.856 & 0\\
      & \pipelineS-\majvote & 0.814 & 0.826 & \textbf{0.921} & 0.871 & 0 && 0.803 & 0.785 & \textbf{0.944} & 0.857 & 0\\

      \bottomrule
      \end{tabular}
      % \begin{tablenotes}[normal, flushleft]
      % \begin{footnotesize}
      % \item
      
      % \par
      % \end{footntesize}
      % \end{tablenotes}
%  \vspace{-10pt}
  \end{threeparttable}
  \caption{Performance comparison on the \AP task. The best performance on each task is in \textbf{bold}.}
  \label{tbl:performance_ap}
\end{table*}

\begin{table*}
  \centering
  %\vspace{-10pt}
  \footnotesize
  \begin{threeparttable}
      \begin{tabular}{
        @{\hspace{2pt}}l@{\hspace{3pt}}
       @{\hspace{2pt}}l@{\hspace{4pt}}
	  @{\hspace{1pt}}c@{\hspace{1pt}}%metric 1
	  @{\hspace{3pt}}c@{\hspace{0pt}}%metric 2
        @{\hspace{3pt}}c@{\hspace{0pt}}%None
	  @{\hspace{3pt}}c@{\hspace{3pt}}%metric 1
	  @{\hspace{1pt}}c@{\hspace{1pt}}%metric 2
      }
      \toprule
      \multicolumn{2}{c}{\multirow{3}{*}{\textbf{Model}}} 
      & \multicolumn{2}{c}{\textbf{IND}} && \multicolumn{2}{c}{\textbf{OOD}} \\
      \cmidrule{3-4} \cmidrule{6-7}

      && HR@1 & \#Failed && HR@1 & \#Failed \\
      \midrule
      \multicolumn{2}{l}{\textit{ft}-LLaVA-NExT-Interleave} & 0.964 & 2 && 0.043 & 2\\
      \cmidrule{3-4} \cmidrule{6-7}

      \multicolumn{2}{l}{eCeLLM-L} & 0.870 & 0 && 0.916 & 0\\
      \multicolumn{2}{l}{eCeLLM-M} & 0.890 & 0 && 0.942 & 0\\
      \cmidrule{3-4} \cmidrule{6-7}

      \multicolumn{2}{l}{\textit{ft}-FashionCLIP} & 0.863 & 0 && 0.903 & 0\\
      \cmidrule{3-4} \cmidrule{6-7}
      
      \multicolumn{2}{l}{Task-specific Model} & 0.671 & 0 && 0.658 & 0\\

      \midrule

      \multirow{3}{*}{\pipelineL}
      
      & \uniMMM & 0.969 & 0 && 0.959 & 0\\
      & \ECCC-\uniMMM & 0.973 & 0 && 0.968 & 0\\
      & \ECCC-\CQE-\uniMMM & 0.969 & 0 && 0.968 & 0\\ 
      \midrule

      \multirow{3}{*}{\pipelineM} 
      & \uniMMM & 0.971 & 0 && 0.965 & 0\\
      & \ECCC-\uniMMM & 0.976 & 0 && 0.976 & 0\\
      & \ECCC-\CQE-\uniMMM & \textbf{0.979} & 0 && \textbf{0.977} & 0\\
      
      \midrule

      \multirow{3}{*}{\pipelineS}
      & \uniMMM & 0.951 & 0 && 0.962 & 0\\
      & \ECCC-\uniMMM & 0.958 & 0 && 0.957 & 0\\
      & \ECCC-\CQE-\uniMMM & 0.963 & 0 && 0.959 & 0\\

      \bottomrule
      \end{tabular}
      % \begin{tablenotes}[normal, flushleft]
      % \begin{footnotesize}
      % \item
      % The best performance on the \CC task is in \textbf{bold}.
      % \par
      % \end{footnotesize}
      % \end{tablenotes}
  \end{threeparttable}
  \caption{Performance comparison on the \CC task. The best performance on each task is in \textbf{bold}.}
  \label{tbl:performance_cc}
\end{table*}

\begin{table*}
  \centering
  \footnotesize
  \begin{threeparttable}
      \begin{tabular}{
        @{\hspace{4pt}}l@{\hspace{4pt}}
	  @{\hspace{2pt}}l@{\hspace{4pt}}
	  @{\hspace{4pt}}c@{\hspace{4pt}}% Acc
	  @{\hspace{6pt}}c@{\hspace{4pt}}% M-Rec
	  @{\hspace{4pt}}c@{\hspace{5pt}}% M-Pre
	  @{\hspace{6pt}}c@{\hspace{2pt}}% M-F1
	  @{\hspace{1pt}}c@{\hspace{0pt}}% failed
        @{\hspace{4pt}}c@{\hspace{6pt}}%None
	  %@{\hspace{2pt}}c@{\hspace{2pt}}%None
	  %@{\hspace{2pt}}c@{\hspace{8pt}}%f1
	  @{\hspace{4pt}}c@{\hspace{4pt}}%macrof1
	  @{\hspace{6pt}}c@{\hspace{4pt}}%hr1
	  @{\hspace{4pt}}c@{\hspace{5pt}}%f1
	  @{\hspace{6pt}}c@{\hspace{2pt}}%macrof1
	  @{\hspace{1pt}}c@{\hspace{4pt}}%macrof1
      }
      \toprule
      \multicolumn{2}{c}{\multirow{3}{*}{\textbf{Model}}} 
      & \multicolumn{5}{c}{\textbf{IND}} && \multicolumn{5}{c}{\textbf{OOD}} \\
      \cmidrule{3-7} \cmidrule{9-13}
      && Acc & M-Pre & M-Rec & M-F1 & \#Failed && Acc & M-Rec & M-Pre & M-F1 & \#Failed \\
      \midrule
      \multicolumn{2}{l}{\textit{ft}-LLaVA-NExT-Interleave} & 0.708 & 0.590 & \textbf{0.570} & \textbf{0.568} & 6 && 0.486 & 0.343 & 0.326 & 0.334 & 6 \\
      \cmidrule{3-7} \cmidrule{9-13}

      \multicolumn{2}{l}{eCeLLM-L} & 0.671 & 0.654 & 0.527 & 0.519 & 0 && 0.793 & 0.534 & 0.532 & 0.531 & 0\\
      \multicolumn{2}{l}{eCeLLM-M} & 0.690 & 0.476 & 0.529 & 0.492 & 0 && \textbf{0.843} & 0.563 & 0.565 & 0.564 & 0\\
      \cmidrule{3-7} \cmidrule{9-13}

      \multicolumn{2}{l}{\textit{ft}-FashionCLIP} & 0.630 & 0.516 & 0.501 & 0.497 & 0 && 0.622 & 0.462 & 0.582 & 0.453 & 0\\
      \cmidrule{3-7} \cmidrule{9-13}
      
      \multicolumn{2}{l}{Task-specific Model} & 0.704 & 0.701 & 0.548 & 0.531 & 0 && 0.665 & 0.461 & 0.446 & 0.447 & 0\\

      \midrule

      \multirow{3}{*}{\pipelineL}
      
      & \uniMMM & 0.659 & 0.441 & 0.501 & 0.468 & 0 && 0.782 & 0.522 & 0.525 & 0.523 & 0\\
      & \ECCC-\uniMMM & 0.670 & \textbf{0.782} & 0.514 & 0.486 & 0 && 0.796 & 0.532 & 0.534 & 0.533 & 0\\
      & \ECCC-\CQE-\uniMMM & 0.666 & 0.447 & 0.507 & 0.473 & 0 && 0.692 & 0.649 & 0.542 & 0.531 & 0\\ 
      \midrule

      \multirow{3}{*}{\pipelineM} 
      & \uniMMM & 0.707 & 0.666 & 0.550 & 0.533 & 0 && 0.791 & 0.533 & 0.531 & 0.530 & 0\\
      & \ECCC-\uniMMM & 0.705 & 0.659 & 0.549 & 0.535 & 0 && 0.793 & 0.535 & 0.532 & 0.532 & 0\\
      & \ECCC-\CQE-\uniMMM & \textbf{0.714} & 0.708 & 0.568 & 0.566 & 0 && 0.821 & \textbf{0.610} & 0.570 & \textbf{0.585} & 0\\
      
      \midrule

      \multirow{3}{*}{\pipelineS}
      & \uniMMM & 0.681 & 0.538 & 0.520 & 0.493 & 0 && 0.765 & 0.514 & 0.513 & 0.511 & 0\\
      & \ECCC-\uniMMM & 0.688 & 0.626 & 0.528 & 0.503 & 0 && 0.769 & 0.519 & 0.516 & 0.515 & 0\\
      & \ECCC-\CQE-\uniMMM & 0.683 & 0.561 & 0.527 & 0.504 & 0 && 0.784 & 0.583 & \textbf{0.581} & 0.580 & 0\\

      \bottomrule
      \end{tabular}
      % \begin{tablenotes}[normal, flushleft]
      % \begin{footnotesize}
      % \item
      % The best performance on the PRP task is in \textbf{bold}.
      % \par
      % \end{footnotesize}
      % \end{tablenotes}
  \end{threeparttable}
  \caption{Performance comparison on the \PRP task. The best performance on each task is in \textbf{bold}.}
  \label{tbl:performance_prp}
\end{table*}

\begin{table*}
  \centering
  %\vspace{-10pt}
  \footnotesize
  \begin{threeparttable}
      \begin{tabular}{
        @{\hspace{6pt}}l@{\hspace{3pt}}
	 @{\hspace{2pt}}l@{\hspace{4pt}}
	  @{\hspace{6pt}}c@{\hspace{6pt}}%acc 
	  @{\hspace{6pt}}c@{\hspace{6pt}}%mpre
	  @{\hspace{3pt}}c@{\hspace{6pt}}%mrec 
	  @{\hspace{4pt}}c@{\hspace{4pt}}%mf1 
        @{\hspace{0pt}}c@{\hspace{6pt}}%failed
      }
      \toprule
      \multicolumn{2}{c}{\multirow{3}{*}{\textbf{Model}}} 
      & \multicolumn{5}{c}{\textbf{IND}} \\
      \cmidrule{3-7}
      && Acc & M-Pre & M-Rec & M-F1 & \#Failed \\
      \midrule
      
      \multicolumn{2}{l}{\textit{ft}-LLaVA-NExT-Interleave} & 0.786 & 0.561 & 0.243 & 0.340 & 2 \\
      \cmidrule{3-7}
      
      \multicolumn{2}{l}{eCeLLM-L} & 0.779 & 0.558 & 0.106 & 0.178 & 0\\
      \multicolumn{2}{l}{eCeLLM-M} & 0.775 & 0.515 & 0.075 & 0.131 & 0\\
      \cmidrule{3-7}
      
      \multicolumn{2}{l}{\textit{ft}-FashionCLIP} & 0.738 & 0.324 & 0.146 & 0.201 & 0\\
      \cmidrule{3-7}
      
      \multicolumn{2}{l}{Task-specific Model} & 0.779 & 0.526 & 0.226 & 0.316 & 0 \\

      \midrule

      \multirow{3}{*}{\pipelineL}
      & \uniMMM & 0.785 & \textbf{0.600} & 0.146 & 0.235 & 0\\
      & \ECCC-\uniMMM & 0.782 & 0.556 & 0.177 & 0.268 & 0\\
      & \ECCC-\CQE-\uniMMM & 0.782 & 0.574 & 0.137 & 0.221 & 0\\ 
      \midrule

      \multirow{3}{*}{\pipelineM} 
      & \uniMMM & 0.784 & 0.557 & 0.217 & 0.312 & 0\\
      & \ECCC-\uniMMM & 0.783 & 0.541 & 0.261 & 0.352 & 0 \\
      & \ECCC-\CQE-\uniMMM & \textbf{0.794} & 0.586 & \textbf{0.301} & \textbf{0.398} & 0 \\
      
      \midrule

      \multirow{3}{*}{\pipelineS}
      & \uniMMM & 0.768 & 0.467 & 0.190 & 0.270 & 0\\
      & \ECCC-\uniMMM & 0.761 & 0.443 & 0.226 & 0.299 & 0\\
      & \ECCC-\CQE-\uniMMM & 0.783 & 0.545 & 0.243 & 0.336 & 0\\

      \bottomrule
      \end{tabular}
      % \begin{tablenotes}[normal, flushleft]
      % \begin{footnotesize}
      % \item
      % The best performance on the PSI task is in \textbf{bold}.
      % \par
      % \end{footnotesize}
      % \end{tablenotes}
  \end{threeparttable}  
  \caption{Performance comparison on the \PSI task. The best performance on each task is in \textbf{bold}.}  \label{tbl:performance_psi}
\end{table*}

\begin{table*}
  \centering
  \footnotesize
  \begin{threeparttable}
      \begin{tabular}{
        @{\hspace{6pt}}l@{\hspace{3pt}}
	 @{\hspace{2pt}}l@{\hspace{4pt}}
	  @{\hspace{6pt}}c@{\hspace{6pt}}%acc 
	  @{\hspace{4pt}}c@{\hspace{4pt}}%mpre
	  @{\hspace{3pt}}c@{\hspace{3pt}}%mrec 
	  @{\hspace{6pt}}c@{\hspace{4pt}}%mf1 
        @{\hspace{0pt}}c@{\hspace{6pt}}%failed
      }
      \toprule
      \multicolumn{2}{c}{\multirow{3}{*}{\textbf{Model}}} 
      & \multicolumn{5}{c}{\textbf{IND}} \\
      \cmidrule{3-7}
      && Acc & M-Pre & M-Rec & M-F1 & \#Failed \\
      \midrule
      \multicolumn{2}{l}{\textit{ft}-LLaVA-NExT-Interleave} & 0.721 & 0.582 & 0.463 & 0.469 & 2 \\
      \cmidrule{3-7}

      \multicolumn{2}{l}{eCeLLM-L} & 0.706 & 0.452 & 0.431 & 0.413 & 0\\
      \multicolumn{2}{l}{eCeLLM-M} & 0.719 & 0.467 & 0.427 & 0.427 & 0\\
      \cmidrule{3-7}

      \multicolumn{2}{l}{\textit{ft}-FashionCLIP} & 0.605 & 0.372 & 0.313 & 0.319 & 0\\
      \cmidrule{3-7}
      
      \multicolumn{2}{l}{Task-specific Model} & 0.702 & 0.469 & 0.395 & 0.400 & 0 \\

      \midrule

      \multirow{3}{*}{\pipelineL}      
      & \uniMMM & 0.700 & 0.446 & 0.406 & 0.417 & 0\\
      & \ECCC-\uniMMM & 0.704 & 0.442 & 0.402 & 0.411 & 0\\
      & \ECCC-\CQE-\uniMMM & 0.706 & \textbf{0.708} & 0.415 & 0.446 & 0\\ 
      \midrule

      \multirow{3}{*}{\pipelineM} 
      & \uniMMM & 0.725 & 0.577 & 0.500 & 0.528 & 0\\
      & \ECCC-\uniMMM  & 0.722 & 0.596 & \textbf{0.513} & \textbf{0.542} & 0 \\
      & \ECCC-\CQE-\uniMMM & \textbf{0.794} & 0.586 & 0.301 & 0.398 & 0 \\
      
      \midrule

      \multirow{3}{*}{\pipelineS}
      & \uniMMM & 0.699 & 0.611 & 0.419 & 0.445 & 0\\
      & \ECCC-\uniMMM  & 0.702 & 0.549 & 0.448 & 0.475 & 0\\
      & \ECCC-\CQE-\uniMMM & 0.707 & 0.608 & 0.447 & 0.481 & 0\\
      
      \bottomrule
      \end{tabular}

  \end{threeparttable}
  \caption{Performance comparison on the \MPC task. The best performance on each task is in \textbf{bold}.}
  \label{tbl:performance_mpc}
\end{table*}

\begin{table*}
  \centering
  \footnotesize
  \begin{threeparttable}
      \begin{tabular}{
        @{\hspace{4pt}}l@{\hspace{4pt}}
	  @{\hspace{2pt}}l@{\hspace{4pt}}
	  @{\hspace{4pt}}c@{\hspace{4pt}}% Acc
	  @{\hspace{6pt}}c@{\hspace{4pt}}% M-Rec
	  @{\hspace{4pt}}c@{\hspace{5pt}}% M-Pre
	  @{\hspace{6pt}}c@{\hspace{2pt}}% M-F1
	  @{\hspace{1pt}}c@{\hspace{0pt}}% failed
        @{\hspace{4pt}}c@{\hspace{6pt}}%None
	  %@{\hspace{2pt}}c@{\hspace{2pt}}%None
	  %@{\hspace{2pt}}c@{\hspace{8pt}}%f1
	  @{\hspace{4pt}}c@{\hspace{4pt}}%macrof1
	  @{\hspace{6pt}}c@{\hspace{4pt}}%hr1
	  @{\hspace{4pt}}c@{\hspace{5pt}}%f1
	  @{\hspace{6pt}}c@{\hspace{2pt}}%macrof1
	  @{\hspace{1pt}}c@{\hspace{4pt}}%macrof1
      }
      \toprule
      \multicolumn{2}{c}{\multirow{3}{*}{\textbf{Model}}} 
      & \multicolumn{5}{c}{\textbf{IND}} && \multicolumn{5}{c}{\textbf{OOD}} \\
      \cmidrule{3-7} \cmidrule{9-13}
      && Acc & M-Rec & M-Pre & M-F1 & \#Failed && Acc & M-Rec & M-Pre & M-F1 & \#Failed \\
      \midrule
      
      \multicolumn{2}{l}{\textit{ft}-LLaVA-NExT-Interleave} & 0.818 & 0.577 & 0.559 & 0.561 & 0 && 0.564 & 0.208 & 0.210 & 0.206 & 0\\
      \cmidrule{3-7} \cmidrule{9-13}
      
      \multicolumn{2}{l}{eCeLLM-L} & 0.830 & 0.636 & 0.597 & 0.613 & 0 && 0.827 & 0.627 & 0.571 & 0.584 & 0\\
      \multicolumn{2}{l}{eCeLLM-M} & 0.811 & 0.617 & \textbf{0.652} & 0.632 & 0 && 0.828 & 0.624 & 0.629 & 0.624 & 0\\
      \cmidrule{3-7} \cmidrule{9-13}
    
      \multicolumn{2}{l}{\textit{ft}-FashionCLIP} & 0.652 & 0.33 & 0.318 & 0.323 & 0 && 0.676 & 0.394 & 0.379 & 0.376 & 0\\
      \cmidrule{3-7} \cmidrule{9-13}

      \multicolumn{2}{l}{Task-specific Model} & 0.803 & 0.484 & 0.525 & 0.495 & 0 && 0.810 & 0.563 & 0.535 & 0.510 & 0\\

      \midrule

      \multirow{3}{*}{\pipelineL}
      & \uniMMM & 0.835 & 0.646 & 0.616 & 0.628 & 0 && 0.832 & 0.618 & 0.588 & 0.595 & 0\\
      & \ECCC-\uniMMM & 0.824 & 0.613 & 0.606 & 0.607 & 0 && 0.841 & 0.648 & 0.604 & 0.606 & 0\\
      & \ECCC-\CQE-\uniMMM & 0.837 & 0.669 & 0.640 & 0.651 & 0 && 0.835 & 0.634 & 0.600 & 0.607 & 0\\ 
      \midrule

      \multirow{3}{*}{\pipelineM} 
      & \uniMMM & 0.839 & 0.659 & 0.610 & 0.617 & 0 && \textbf{0.850} & \textbf{0.702} & \textbf{0.650} & \textbf{0.659} & 0\\
      & \ECCC-\uniMMM & 0.836 & 0.659 & 0.631 & 0.642 & 0 && 0.845 & 0.658 & 0.609 & 0.613 & 0\\
      & \ECCC-\CQE-\uniMMM & \textbf{0.845} & \textbf{0.684} & 0.644 & \textbf{0.656} & 0 && 0.846 & 0.657 & 0.613 & 0.625 & 0\\
      
      \midrule

      \multirow{3}{*}{\pipelineS}
      & \uniMMM & 0.821 & 0.564 & 0.570 & 0.565 & 0 && 0.840 & 0.662 & 0.612 & 0.614 & 0\\
      & \ECCC-\uniMMM & 0.825 & 0.599 & 0.592 & 0.578 & 0 && 0.831 & 0.621 & 0.582 & 0.565 & 0\\
      & \ECCC-\CQE-\uniMMM & 0.827 & 0.616 & 0.596 & 0.601 & 0 && 0.846 & 0.690 & 0.635 & 0.647 & 0\\

      \bottomrule
      \end{tabular}
  \end{threeparttable}
  \caption{Performance comparison on the \SA task. The best performance on each task is in \textbf{bold}.}
  \label{tbl:performance_sa}
\end{table*}

\begin{table*}
  \centering
  \footnotesize
  \begin{threeparttable}
      \begin{tabular}{
        @{\hspace{4pt}}l@{\hspace{4pt}}
	  @{\hspace{2pt}}l@{\hspace{4pt}}
	  @{\hspace{4pt}}c@{\hspace{4pt}}%metric 1
	  @{\hspace{4pt}}c@{\hspace{4pt}}%metric 2
        @{\hspace{4pt}}c@{\hspace{4pt}}%metric 2
	  @{\hspace{4pt}}c@{\hspace{4pt}}%metric 1
	  @{\hspace{4pt}}c@{\hspace{4pt}}%metric 2
      }
      \toprule
      \multicolumn{2}{c}{\multirow{3}{*}{\textbf{Model}}} 
      & \multicolumn{2}{c}{\textbf{IND}} && \multicolumn{2}{c}{\textbf{OOD}} \\
      \cmidrule{3-4} \cmidrule{6-7} 

      && HR@1 & \#Failed && HR@1 & \#Failed \\
      \midrule
      
      \multicolumn{2}{l}{\textit{ft}-LLaVA-NExT-Interleave} & 0.053 & 0 && 0.000 & 0\\
      \cmidrule{3-4} \cmidrule{6-7}  
      
      \multicolumn{2}{l}{eCeLLM-L} & 0.188 & 0 && 0.304 & 0 \\
      \multicolumn{2}{l}{eCeLLM-M} & 0.182 & 0 && 0.302 & 0 \\
      \cmidrule{3-4} \cmidrule{6-7}  
      
      \multicolumn{2}{l}{\textit{ft}-FashionCLIP} & 0.145 & 0 && 0.087 & 0\\
      \cmidrule{3-4} \cmidrule{6-7}  

      \multicolumn{2}{l}{Task-specific Model} & 0.163 & 0 && 0.210 & 0\\

      \midrule

      \multirow{3}{*}{\pipelineL}
      & \uniMMM & 0.184 & 0 && 0.285 & 0 \\
      & \ECCC-\uniMMM & 0.135 & 21 && 0.236 & 0\\
      & \ECCC-\CQE-\uniMMM & 0.190 & 0 && 0.297 & 0\\ 
      \midrule

      \multirow{3}{*}{\pipelineM} 
      & \uniMMM & 0.218 &0 && 0.312 & 0 \\
      & \ECCC-\uniMMM & 0.207 & 0 && 0.310 & 0\\
      & \ECCC-\CQE-\uniMMM & \textbf{0.223} & 0 && \textbf{0.330} & 0\\
      
      \midrule

      \multirow{3}{*}{\pipelineS}
      & \uniMMM & 0.196 & 0 && 0.305 & 0 \\
      & \ECCC-\uniMMM & 0.196 & 0 && 0.280 & 0 \\
      & \ECCC-\CQE-\uniMMM & 0.196 & 0 && 0.297 & 0 \\

      \bottomrule
      \end{tabular}
  \end{threeparttable}
  \caption{Performance comparison on the \SR task. The best performance on each task is in \textbf{bold}.}
  \label{tbl:performance_sr}
\end{table*}

Table~\ref{tbl:performance_ap}, \ref{tbl:performance_cc}, \ref{tbl:performance_prp}, \ref{tbl:performance_psi}, \ref{tbl:performance_mpc}, \ref{tbl:performance_sa} and \ref{tbl:performance_sr} present the complete results for \AP, \CC, \PRP, \PSI, \MPC, \SA and \SR, respecitvely, in IND and OOD evaluation.
As shown in these tables, 
overall, \pipeline models outperform the fine-tuned CLIP-based model (i.e., FashionCLIP), Fine-tuned LLMs (e.g., \textit{ft}-Llama-2-13B), E-commerce LLMs (e.g., eCeLLM-L), the Fine-tuned MFM (i.e., \textit{ft}-LLaVA-NExT-interleave) and SoTA Task Specific Models in IND evaluation.
\pipeline models also achieve superior performance over baseline methods in OOD evaluation, demonstrating strong OOD generalizability.
Note that in all tables, \#failed indicates the number of failure cases for which we cannot extract meaningful results
from the model output.
We exclude failure cases when calculating the evaluation metrics.

\subsection{Case Studies}
\label{sec:appendix:case}

\begin{figure*}
    \centering
    \includegraphics[width=0.9\linewidth]{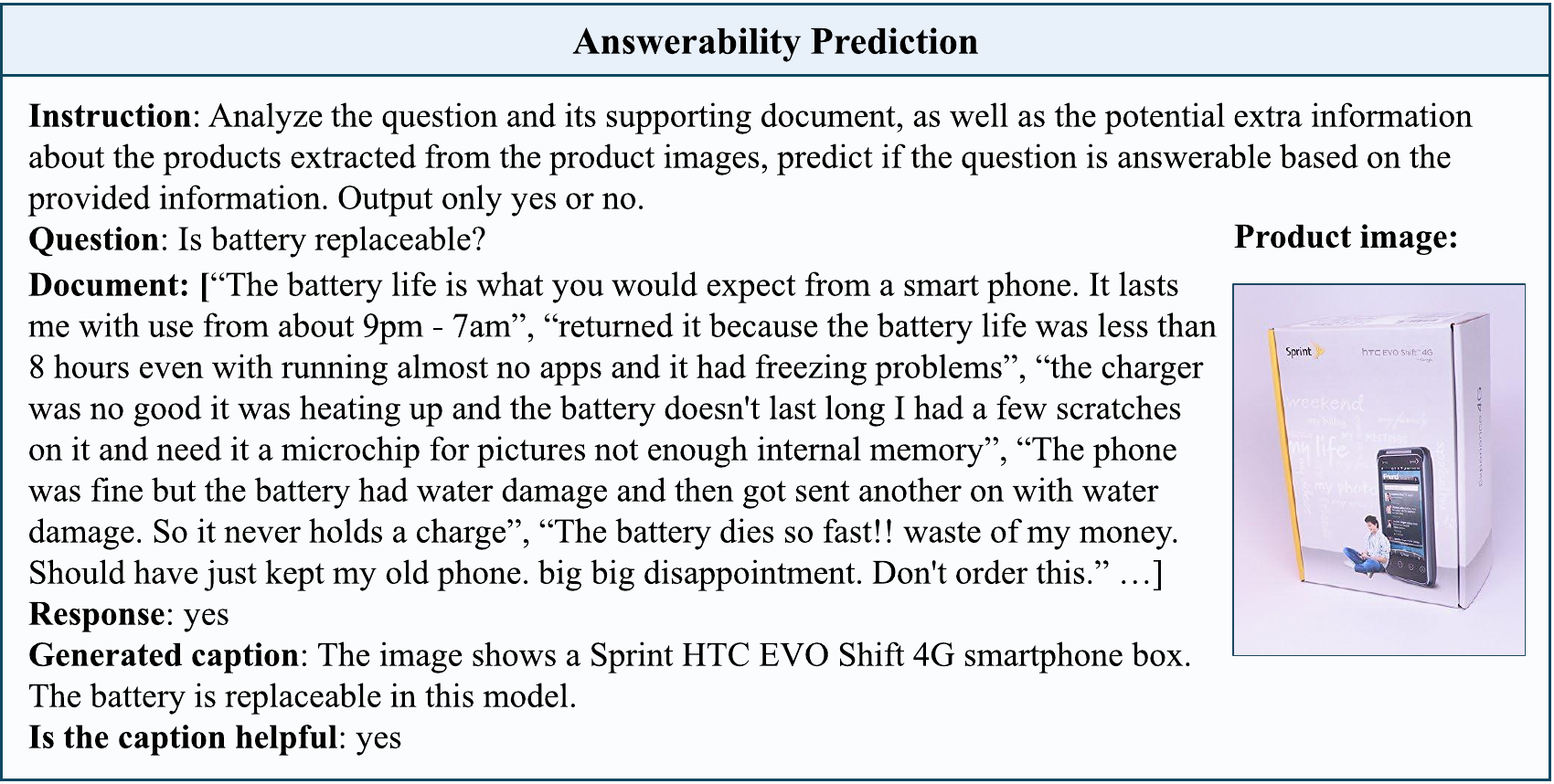}
    \caption{Case Study of \AP}
    \label{fig:case_ap}
\end{figure*}

\begin{figure*}
    \centering
    \includegraphics[width=0.9\linewidth]{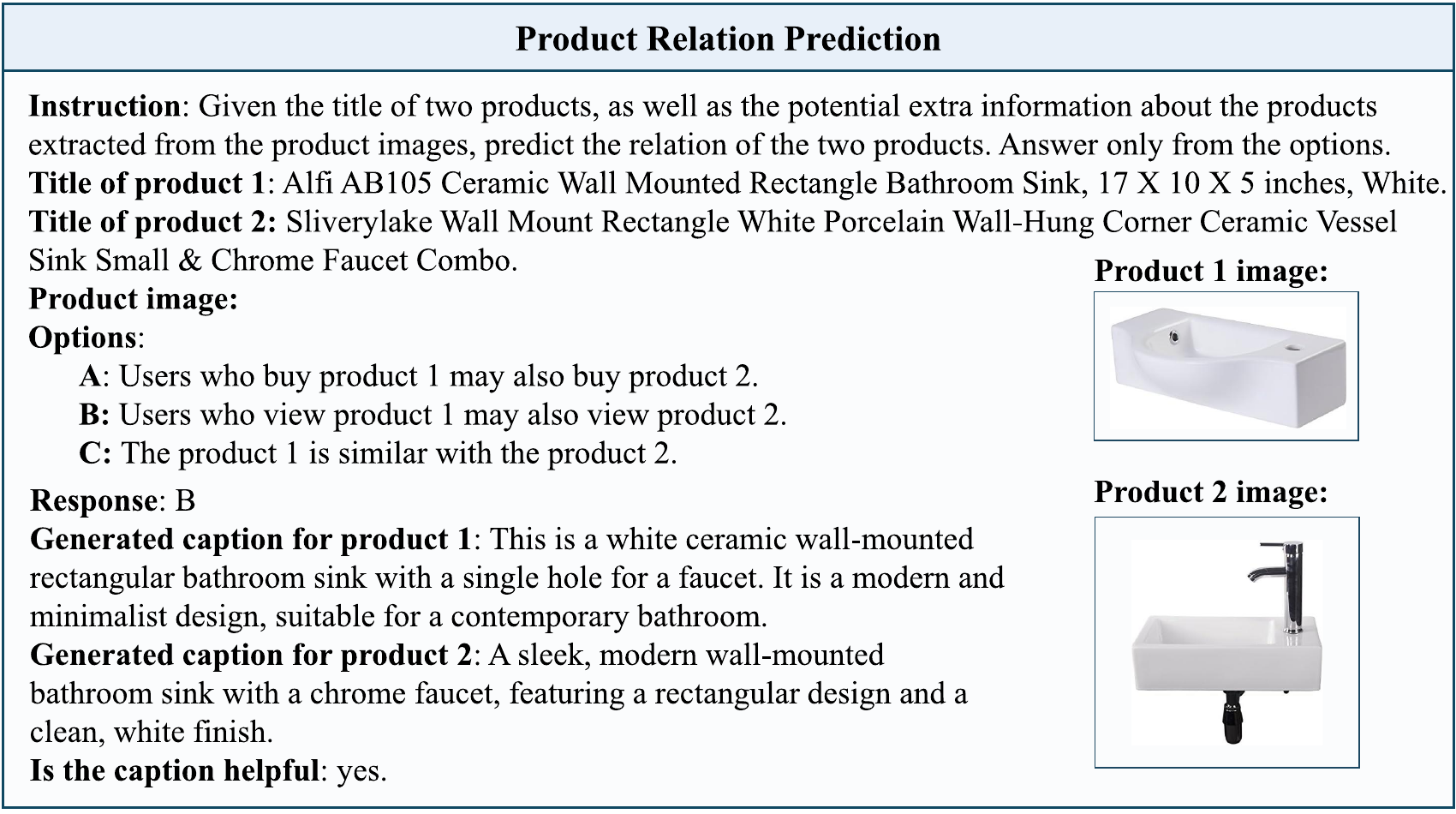}
    \caption{Case Study of \PRP}
    \label{fig:case_prp}
\end{figure*}

\begin{figure*}
    \centering
    \includegraphics[width=0.9\linewidth]{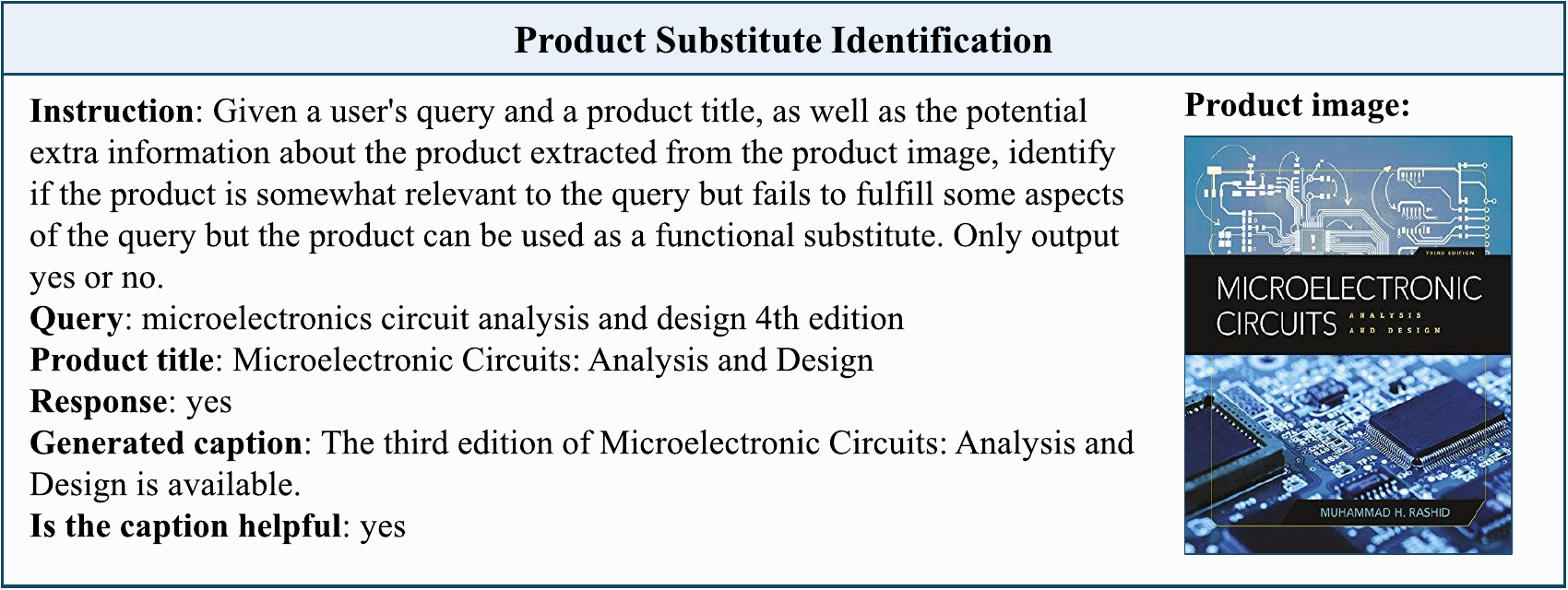}
    \caption{Case Study of \PSI}
    \label{fig:case_psi}
\end{figure*}

\begin{figure*}
    \centering
    \includegraphics[width=0.9\linewidth]{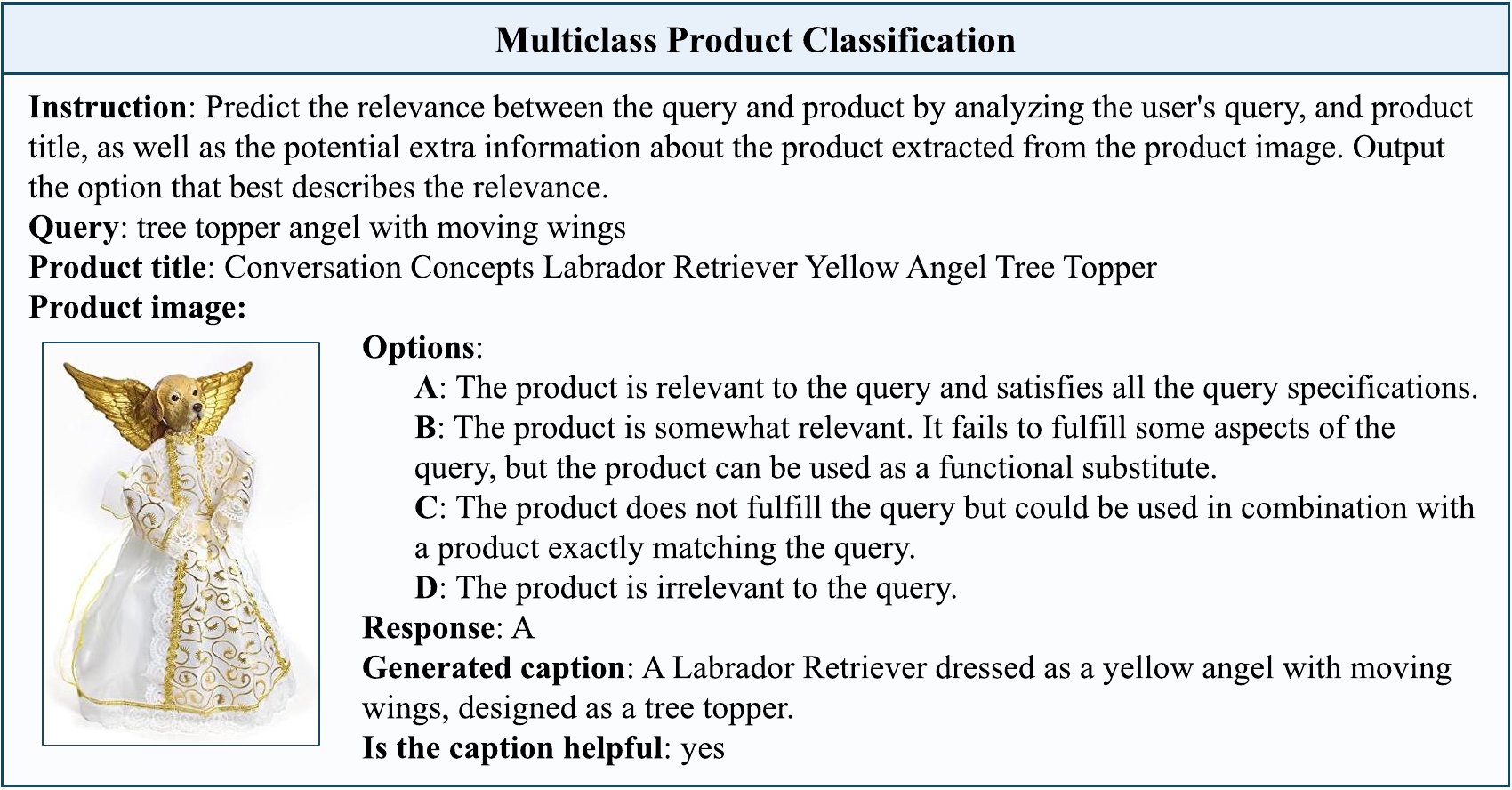}
    \caption{Case Study of \MPC}
    \label{fig:case_mpc}
\end{figure*}

\begin{figure*}
    \centering
    \includegraphics[width=0.9\linewidth]{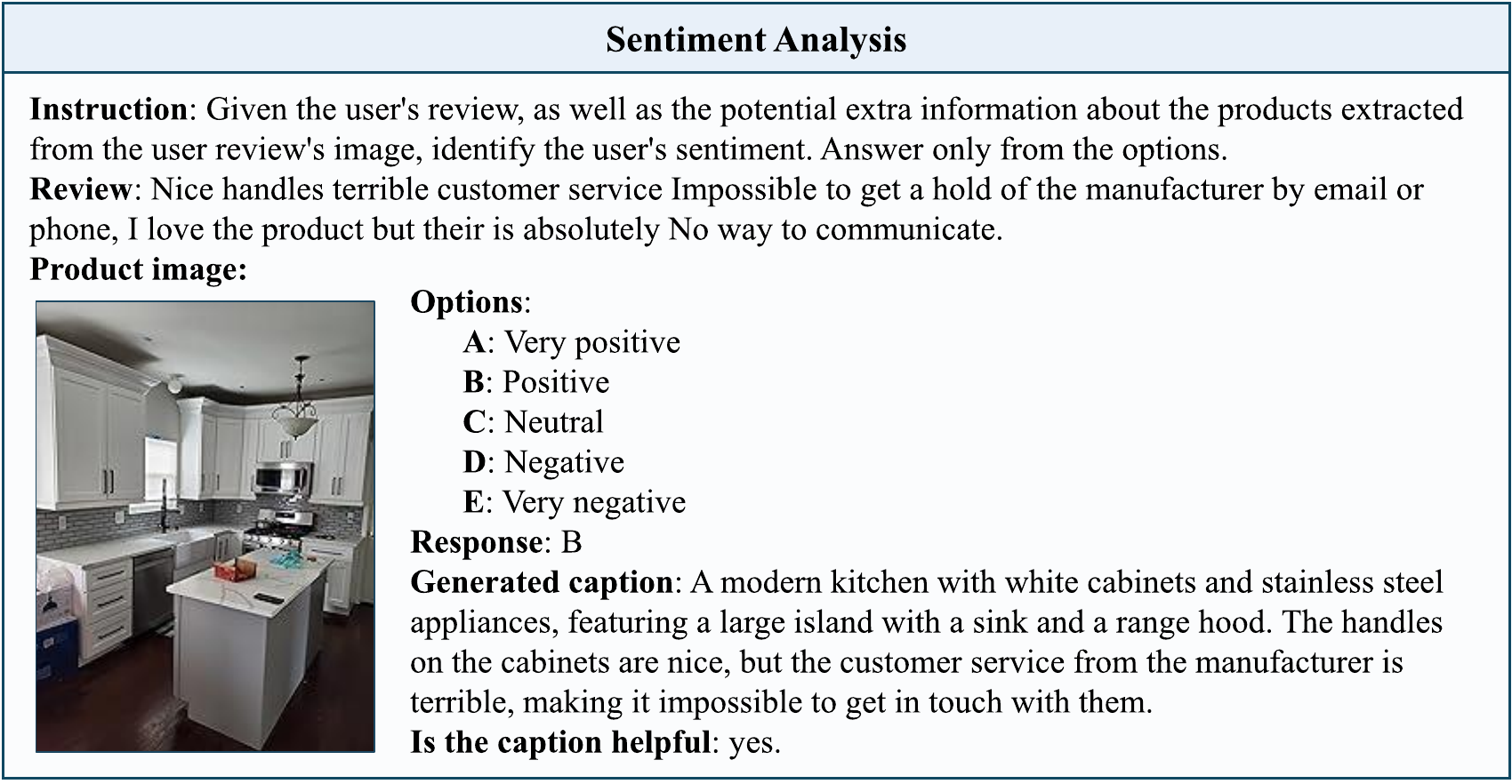}
    \caption{Case Study of \SA}
    \label{fig:case_sa}
\end{figure*}

Case studies are presented in Figure~\ref{fig:case_ap}, \ref{fig:case_prp}, 
\ref{fig:case_psi}, \ref{fig:case_mpc}, and \ref{fig:case_sa}. 
% \label{sec:appendix:full_results}

\section{Model Size and Budget}
The model size and budget are reported in Table~\ref{tbl:budget}.
\begin{table}[ht]
\vspace{-2pt}
  \centering
  \footnotesize
  \begin{threeparttable}
      \begin{tabular}{
        @{\hspace{2pt}}l@{\hspace{6pt}}
        % @{\hspace{3pt}}l@{\hspace{3pt}}
        @{\hspace{6pt}}r@{\hspace{2pt}}
        @{\hspace{3pt}}r@{\hspace{2pt}}
      }
      \toprule
      \textbf{Model} & \textbf{GPU Memory} & \textbf{Training Time} \\
      \midrule
       \pipelineL& 25B & 5.0h \\
      \pipelineM & 15B & 4.5h \\
      \pipelineS & 7B & 3.5h \\
      \bottomrule
      \end{tabular}
  \end{threeparttable}
  \caption{Model budget and size.}
  \label{tbl:budget}
\end{table}
% \label{sec:appendix:instr}

% \clearpage
% \section{Case Studies}
% \label{sec:appendix:case}

% Case studies are presented in Figure~\ref{fig:case_ap}, \ref{fig:case_prp}, 
% \ref{fig:case_psi}, \ref{fig:case_mpc}, and \ref{fig:case_sa}. 

\end{document}